\documentclass[preprint,12pt]{elsarticle}

\usepackage{amssymb}

\usepackage{times}
\usepackage{epsfig}
\usepackage{graphicx}
\usepackage{amsmath}
\usepackage{amssymb}
\usepackage{algorithm}
\usepackage{algorithmic}
\usepackage{soul, color}
\usepackage{multicol}
\usepackage{bm}
\usepackage{color, colortbl}
\usepackage{multirow}
\usepackage[table]{xcolor}
\usepackage{url}

\long\def\comment#1{}
\definecolor{Gray}{gray}{0.8}
\newcolumntype{g}{>{\columncolor{Gray}}c}

\journal{Pattern Recognition}

\begin{document}

\begin{frontmatter}



\title{Efficient Estimation of the number of neighbours \\ in Probabilistic K Nearest Neighbour Classification}

\author[label1]{Ji~Won~Yoon}
\author[label2]{Nial~Friel}

\address[label1]{Center for information security technology (CIST), Korea University, Korea}
\address[label2]{School of Mathematical Sciences, University College Dublin, Ireland}




\begin{abstract}
Probabilistic k-nearest neighbour (PKNN) classification has been introduced to improve the performance of original k-nearest neighbour (KNN) classification algorithm by explicitly modelling uncertainty in the classification of each feature vector. However, an issue common to both KNN and PKNN is to select the optimal number of neighbours, $k$. The contribution of this paper is to incorporate the uncertainty in $k$ into the decision making, and in so doing use Bayesian model averaging to provide improved classification. Indeed the problem of assessing the uncertainty in $k$ can be viewed as one of statistical model selection which is one of the most important technical issues in the statistics and machine learning domain. In this paper, a new functional approximation algorithm is proposed to reconstruct the density of the model (order) without relying on time consuming Monte Carlo simulations. In addition, this algorithm avoids cross validation by adopting Bayesian framework. The performance of this algorithm yielded very good performance on several real experimental datasets.

\end{abstract}


\end{frontmatter}



\section{Introduction}

Supervised classification is a very well studied problem in the machine learning and statistics literature, where the 
$k-$nearest neighbour algorithm (KNN) is one of the most popular approaches. It amounts to assigning an unlabelled 
class to the most common class label among $k$ neighbouring feature vectors. One of the key issues in implementing
this algorithm is choosing the number of neighbours $k$, and various flavours of cross validation are used for this purpose. 
However a drawback to kNN is that it does not have a probabilistic interpretation, for example, no uncertainty
is associated with the inferred class label. 

There have been several recent papers which addressed this deficiency, 
\cite{Holmes02:PKNN,cuc:mar09,Manocha07:PKNN,fri:pet11}. Indeed from such a Bayesian perspective the issue of 
choosing the value of $k$ can be viewed as a model (order) selection problem. To date, there exist several different 
approaches to tackle the model selection problem. One of the most popular approaches is based on \emph{information criteria} 
including the Akaike Information Criterion (AIC), the Schwarz's Bayesian Information Criterion (BIC) and the Deviance 
Information Criterion (DIC) \cite{Akaike74:AIC,Schwarz78:BIC,Spiegelhalter02:bayesianmeasures}. Given a particular model 
${\cal M}_{k}$, the well-known AIC and BIC are defined by 
$AIC ({\cal M}_{k}) = -2\log L({\cal M}_{k}) + 2e({\cal M}_{k})$ and 
$BIC ({\cal M}_{k}) = -2\log L({\cal M}_{k}) + e({\cal M}_{k})\log N$ for $N$ observations where 
$L({\cal M}_{k})$ and $e({\cal M}_{k})$ denote the likelihood and the number of parameters of ${\cal M}_{k}$, respectively.

It is known that many fast functional approximations or \emph{information criterion} techniques do not adequately approximate the 
underlying posterior distribution of the model order. Furthermore, Monte carlo based estimators can provide approximate 
distributions of the model order, but typically require excessive computation time.

Our main contribution is to propose a new functional approximation technique to infer the posterior distribution of the 
model order, $p(K|\mathcal{Y})$ where $K$ and $\mathcal{Y}$ denote the model order and observations, respectively. In particular, 
this paper demonstrates the applicability of the proposed algorithm by addressing the problem of finding the number of neighbours 
$k$ for probabilistic k-Nearest Neighbour (PKNN) classification. In addition, we designed a new symmetrized neighbouring structure 
for the KNN classifier in order to conduct a fair comparison. From an application point of view, we also classified several 
benchmark datasets and a few real experimental datasets using the proposed algorithms.

In addition to model selection, we also consider improvements of the KNN approach itself for the purpose of a fair 
comparison. Although conventional KNN based on euclidean distance is widely used in many application domains, the conventional 
KNN is not a correct model in that it does not guarantee the symmetricity of the neighbouring structure.

It is important to state that PKNN formally defines a Markov random field over the joint distribution of the class labels.
In turn this yields a complication from an inferential point of view, since it is well understood that the Markov random field
corresponding to likelihood
of the class labels involves an intractable normalising constant, sometimes called the partition function in statistical physics,
rendering exact calculation of the likelihood function almost always impossible.
Inference for such complicated likelihoods function is an active field of research. In the context of PKNN \cite{Holmes02:PKNN}
and \cite{Manocha07:PKNN} use the pseudo-likelihood function \cite{bes74} as an approximation to the true likelihood.
While \cite{cuc:mar09} and \cite{fri:pet11} consider improvements to pseudolikelihood by using a Monte Carlo auxiliary variable 
technique, the exchange algorithm, \cite{Murray06} which targets the posterior distribution which involves the true intractable 
likelihood function.
Bayesian model selection is generally a computationally demanding exercise, particularly in the current context, due to the
intractability of the likelihood function, and for this reason we use a pseudolikelihood approximation throughout this paper, 
although our efforts are focused on efficient means to improve upon this aspect using composite likelihood approximations
\cite{var:reid:fir11}.

This paper consists of several sections. Section \ref{section: Statistical Background} includes the background of the statistical 
approaches used in this paper. This section shows two main techniques, k-Nearest Neighbour (KNN) classification and Integrated 
Laplace Approximation (INLA). For the extended literature review, probabilistic kNN (PKNN) is explained with details. The 
proposed algorithm is introduced in the section \ref{section: Proposed Approach}. In this section, we introduce a generic 
algorithm to reconstruct and approximate the underlying model order posterior $p(K|{\cal Y})$ and to efficiently search for 
the optimal model order $K^{*}$. Afterwards, this section includes how to apply the generic algorithm into PKNN. In section 
\ref{section: Simulation Results}, PKNN adopting the proposed algorithms have applied to several real datasets. Finally, we 
conclude this paper with some discussion of sections \ref{section: Discussion} and \ref{section: Conclusion}.

\section{Related Work}
One of the main aims of this paper is to explore nearest neighbour classification from a model selection perspective. Some
popular model selection approaches in the literature include the following. 
Grenander et al. \cite{Grenander94:JumpProcess,Stephens00:ModelSelection} proposed a model selection 
algorithm which is based on jump-diffusion dynamics with the essential feature that at random times the process jumps between 
parameter spaces in different models and different dimensions. Similarly, Markov birth-death processes and point processes can 
be considered. 
One of the most popular approaches to infer the posterior distribution and to explore model uncertainty is Reversible 
Jump Markov Chain Monte Carlo developed by Richardson and Green \cite{Richardson97:GMM}.
The composite model approach of Carlin and Chib \cite{Carlin95:BayesianModelSelection} is a further approach. The relationships 
between the issue of choice of pseudo-prior in the case of Carlin and Chib's product composite model  and the choice of proposal 
densities in the case of reversible jump are discussed by Godsill \cite{Godsill01:ModelSelection}.

In addition, there are a lot of similarities in the clustering domain. For instance, many clustering algorithms such as K-means 
algorithms, Gaussian Mixture Model (GMM), and Spectral clustering have also the challenging difficulty to infer the number of 
clusters $K$ as similarly shown in the estimation of the number of neighbours $K$ of the (P)KNN. 

\section{Statistical Background}
\label{section: Statistical Background}

\subsection{k-Nearest Neighbour (kNN) model}
In pattern recognition, the k-Nearest Neighbour algorithm (kNN) is one of the most well-known and useful non-parametric methods 
for classifying and clustering objects based on classified features which are close, in some sense, in the feature space. The kNN is designed with the concept that 
labels or classes are determined by a majority vote of its neighbours. However, along with such a simple implementation, the kNN 
has a sensitivity problem from the locality which are generated from two difficult problems: estimating the decision boundary to 
determine the boundary complexity and the number of neighbours to be voted. In order to address this problem, adaptive kNN is 
proposed to  efficiently and effectively calculate the number of neighbours and the boundary 
\cite{Wang06:neighborhoodsize,Hand03:KNN,Domeniconi00adaptivemetric,Guo10:KNN}. In addition, the probabilistic kNN (PKNN) model 
which is more robust than the conventional kNN has been introduced and developed by Markov chain Monte carlo to estimate the number 
of neighbours \cite{Holmes02:PKNN,Manocha07:PKNN}. In this paper, we use the PKNN model since it provides proper 
likelihood term given a particular model with $k$ neighbours.
\begin{figure}[h!]
\centering
\begin{tabular}{cccc}
\includegraphics[scale=0.4]{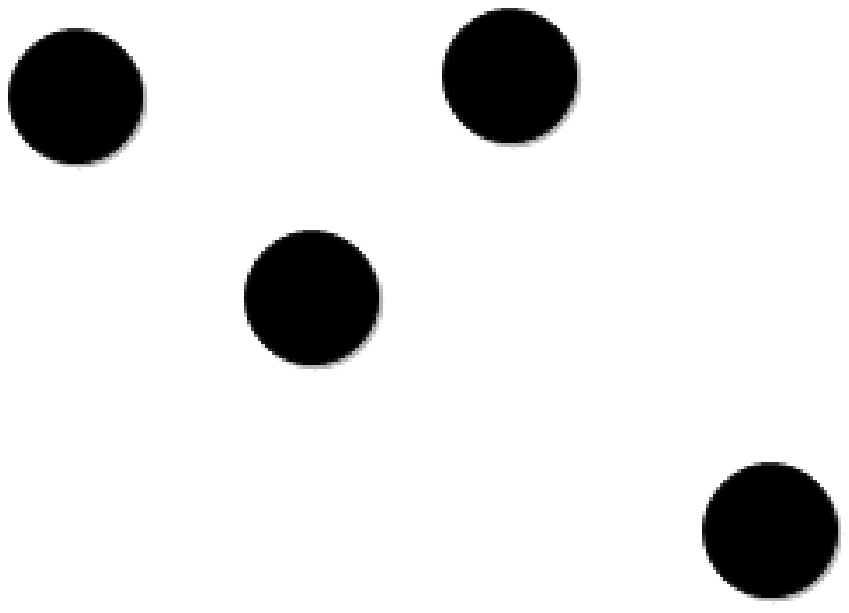}&
\includegraphics[scale=0.4]{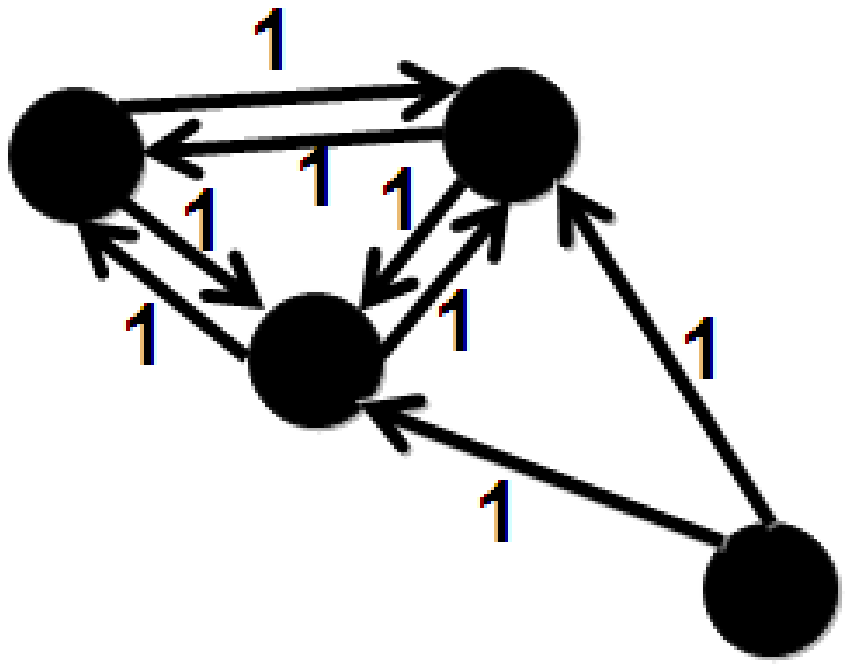}&
\includegraphics[scale=0.4]{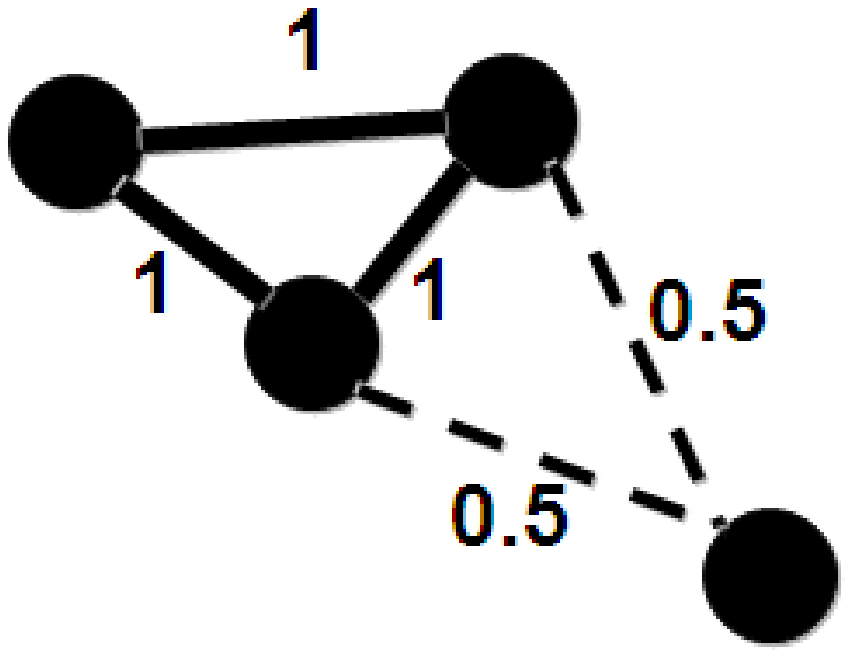}\cr
(a) Given data&
(b) Asymmetric &
(c) Symmetrised \cr
&
PKNN (K=2)&
Boltzmann Model \cr
&
&
for PKNN
\end{tabular}
\caption{Topological Explanation of PKNN}
\label{fig: Topological Explanation of PKNN}
\end{figure}

\subsubsection{An asymmetric Pseudo-likelihood of PKNN}
\label{section: An asymmetric Pseudo-likelihood of PKNN}
Let $\{{(z_{1}, {\bf y}_{1}), (z_{2}, {\bf y}_{2}), \cdots, (z_{N}, {\bf y}_{N})\}}$ where each $z_{i}\in\{1, 2, \cdots, C\}$ denote the class label and $d$ dimensional feature vector ${\bf y}_{i}\in\Re^{d}$. Then, the pseudo-likelihood of the probabilistic kNN (PKNN) proposed by \cite{Holmes02:PKNN} can be formed as
\begin{equation}
p({\bf z}|{\bf y}, \beta, K) \approx \prod_{i=1}^{N} \frac
{\exp \left\{ \frac{\beta}{K} \sum_{j\in ne(i)} \delta_{z_{i}, z_{j}}\right\}}
{\sum_{c\in{\bf C}}\exp \left\{ \frac{\beta}{K} \sum_{j\in ne(i)} \delta_{c, z_{j}} \right\}}
\label{eq: likelihood of PKNN}
\end{equation}
where the unknown scaling value $\beta>0$ and ${\bf C}$ is a set of classes, $K$ denotes the number of neighbours and $\delta_{a, b}=1$ if $a=b$ and $0$ otherwise. In this equation, $ne(\cdot)$ represents the set of neighbours.

Suppose that we have four data points as shown in Fig. \ref{fig: Topological Explanation of PKNN}-(a). Given $K=2$, we have an 
interesting network structure in Fig. \ref{fig: Topological Explanation of PKNN}-(b) from this conventional PKNN. In this 
subgraph, arrows direct the neighbours. As we can see in the Fig. \ref{fig: Topological Explanation of PKNN}-(b), some pairs of 
data points (nodes) are bidirectional but others are unidirectional, resulting in an asymmetric phenomena. Unfortunately, this 
asymmetric property does not satisfy the Markov Random Field assumption which can be implicitly applied in 
Eq. (\ref{eq: likelihood of PKNN}).

\subsubsection{A symmetrised Boltzmann modelling for pseudo-likelihood of PKNN}

Since the pseudo-likelihood of the conventional probabilistic kNN is not symmetrised an approximate symmetrised model has been 
proposed for PKNN \cite{Cucala_Marin_Robert_Titterington_2008} as \newline\newline
\begin{equation}
p({\bf z}|{\bf y}, \beta, K)\approx \prod_{i=1}^{N} \frac
{\exp \left\{ \frac{\beta}{K} (\sum_{j\in ne(i)} \delta_{z_{i}, z_{j}} + \sum_{i\in ne(k)} \delta_{z_{i}, z_{k}})\right\}}
{\sum_{c\in{\bf C}}\exp \left\{ \frac{\beta}{K} (\sum_{j\in ne(i)} \delta_{c, z_{j}}+\sum_{i\in ne(k)} \delta_{z_{i}, z_{k}}) \right\}}.
\label{eq: approximated symmetirc likelihood of PKNN}
\end{equation}
The Boltzmann modeling of PKNN resolves the asymmetric problem which arises from the conventional PKNN of 
Eq. (\ref{eq: likelihood of PKNN}). However, the Boltzmann modeling reconstructs the symmetrised network by averaging the 
asymmetrised effects from the principal structure of PKNN as shown in Fig. \ref{fig: Topological Explanation of PKNN}-(c). This 
brings different interaction rate among the edges. In the subgraph, two edges have a value of a half and all others have a value
of one and so this difference may yield an inaccurate Markov Random Field model again.

\subsection{Estimation of PKNN by Markov chain Monte Carlo (MCMC) - a conventional way}
The most popular approach to estimate the parameters of PKNN is using Markov chain Monte Carlo (MCMC). In this paper, PKNN via 
MCMC is also used for performance comparison. In particular, there are two different version of MCMC. 

The first approach is to infer the unknown model parameters ($\tilde{\beta}$ and $\tilde{K}$) in the training step via MCMC. 
Afterward, given these estimate values, we can classify the new data from the testing set straightforwardly using the conditional 
posterior $p(z_{i}|{\bf y}, {\bf z}, {\bf y}^{'}, \tilde{\beta}, \tilde{K})$. Suppose that we need to reconstruct the target 
posterior $p(\beta, K|{\bf z}, {\bf y})$ given the observations ${\bf z}$ and ${\bf y}$ which is a set of training data. The 
standard MCMC approach uses a Metropolis-Hasting (MH) algorithm, so that each unknown parameter 
is updated according to an acceptance probability
\[
\mathcal{A} = \min \left\{ 1,
\frac{p({\bf z}|{\bf y}, \hat{\beta}, \hat{K})p(\hat{\beta})p(\hat{K})q(\beta, K)}{p({\bf z}|{\bf y}, \beta, K)p(\beta)p(K)q(\hat{\beta}, \hat{K})}
\right\}
\]
where $\hat{\beta}$ and $\hat{K}$ denote the proposed new parameters. In the training step, we estimate $\tilde{\beta}$ and 
$\tilde{K}$ from the above MCMC simulation. Afterwards, we simply classify the testing datasets given $\tilde{\beta}$ and 
$\tilde{K}$. That is, given a testing set we can estimate the classes by
\[
z^{'*}=\arg_{z^{'}}\max p(z^{'}|{\bf y}, {\bf z}, {\bf y}^{'}, \tilde{\beta}, \tilde{K})
\]
for a new test data ${\bf y}^{'}$ and its unknown label $z^{'}$. However, since the uncertainty of the model parameters is ignored 
in the testing step of the first approach, the first approach with two separate steps (training and testing) is less preferred from 
a statistical point of view although it is often used in practice. Unlike the first approach, the second approach jointly 
estimates the hidden model parameters to incorporate this uncertainty while classifying the testing datasets. In the second approach 
we reconstruct not  the conditional distribution $p(z^{'}|{\bf y}, {\bf z}, {\bf y}^{'}, \tilde{\beta}, \tilde{K})$ but a 
marginalized distribution $p(z^{'}|{\bf y}, {\bf z}, {\bf y}^{'})$ by jointly estimating parameters. In this case, the target density is 
not $p(\beta, K|{\bf z}, {\bf y})$ but $p(\beta, K, z^{'}|{\bf z}, {\bf y}, {\bf y}^{'})$. Then each unknown 
parameter from the marginalized density is updated according to the modified acceptance probability
\begin{equation}
\mathcal{A} = \min \left\{ 1,
\frac{p(\hat{z}^{'}, {\bf z}|{\bf y}, {\bf y}_{i}, \hat{\beta}, \hat{K})p(\hat{\beta})p(\hat{K})q(z^{'}, \beta, K)}{p(z^{'}, {\bf z}|{\bf y}, {\bf y}_{i}, \beta, K)p(\beta)p(K)q(\hat{z}^{'}, \hat{\beta}, \hat{K})}
\right\}.
\end{equation}
In this paper, we use the second approach to infer the parameters and classify the data for MCMC simulation for comparison since the joint estimation to obtain the marginalized distribution considers the uncertainty even in the classification of the new dataset. We simply design $q(\hat{z}^{'}, \hat{\beta}, \hat{K})=q(\hat{z}^{'})q(\hat{\beta})q(\hat{K})$ and each proposal distribution is defined by
\begin{eqnarray}
q(\hat{z}^{'}) &=& p(\hat{z}^{'}|{\cal Y}, \hat{\beta}, \hat{K}) =\frac{
p(\hat{z}^{'}, {\bf z}|{\bf y}, {\bf y}_{i}, \hat{\beta}, \hat{K}) 
}{\sum_{c\in{\bf C}}p(\hat{z}^{'}=c, {\bf z}|{\bf y}, {\bf y}_{i}, \hat{\beta}, \hat{K}) }\cr
q(\hat{\beta})&=& \mathcal{N} (\hat{\beta}; \beta, 0.1)\cr
q(\hat{K}) &=& p(\hat{K}) = \frac{1}{K_{\max}}
\end{eqnarray}
where we set $\beta_{a}=2$ and $\beta_{b}=10$ for the Gamma distribution. Given this particular setting of the proposal distribution, we obtain the simplified acceptance probability
\begin{equation}
\mathcal{A} = \min \left\{ 1,
\frac{\sum_{c\in{\bf C}}p(\hat{z}^{'}=c, {\bf z}|{\bf y}, {\bf y}_{i}, \hat{\beta}, \hat{K})p(\hat{\beta})q(\beta)
 }{\
\sum_{s\in{\bf C}}p(z^{'}=s, {\bf z}|{\bf y}, {\bf y}_{i}, \beta, K) p(\beta)q(\hat{\beta})}
\right\}.
\end{equation}

\subsection{Integrated Nested Laplace Approximation (INLA)}

Suppose that we have a set of hidden variables ${\bf f}$ and a set of observations ${\cal Y}$, respectively. MCMC can of course be 
used to infer the marginal density $p({\bf f}|{\bf y})=\int p({\bf f}, \theta|{\bf y})d\theta$  where $\theta$ is a set of control 
parameters. In order to efficiently build the target density, we apply a remarkably fast and accurate functional approximation 
based on the Integrated Nested Laplace Approximation (INLA) developed by \cite{Rue09:INLA}. This algorithm approximates the marginal 
posterior $p({\bf f|{\cal Y}})$ by
\begin{eqnarray}
p({\bf f}|{\cal Y}) &=& \int p({\bf f}|{\cal Y}, \theta)p(\theta|{\cal Y})d\theta \cr
&\approx & \int \tilde{p}({\bf f}|{\cal Y}, \theta)\tilde{p}(\theta|{\cal Y})d\theta\cr
&\approx & \sum_{\theta_{i}}\tilde{p}({\bf f}|{\cal Y}, \theta)\tilde{p}(\theta|{\cal Y})\Delta_{\theta_{i}}
\end{eqnarray}
where
\begin{equation}
\tilde{p}(\theta|{\cal Y}) \propto \left.\frac{p({\bf f}, {\cal Y}, \theta)}{p_{F}({\bf f}|{\cal Y}, \theta)}\right|_{{\bf f}={\bf f}^{*}(\theta)}
= \left.\frac{p({\cal Y}|{\bf f}, \theta)p({\bf f}|\theta)p(\theta)}{p_{F}({\bf f}|{\cal Y}, \theta)}\right|_{{\bf f}={\bf f}^{*}(\theta)}.
\label{eq: marginal posterior of theta in INLA}
\end{equation}
Here, $F$ denotes a simple functional approximation close to $p({\bf f}|{\cal Y}, \theta)$ such as a Gaussian approximation and 
${\bf f}^{*}(\theta)$ is a value of the functional approximation. For the simple Gaussian approximation case, the proper choice 
of ${\bf f}^{*}(\theta)$ is the mode of the Gaussian approximation of $p_{G}({\bf f}|{\cal Y}, \theta)$. Given the log of the posterior, 
we can calculate the mode ${\theta}^{*}$ and its Hessian matrix ${\bf H}_{\theta}^{*}$ via Quasi-Newton style optimization by 
$\theta^{*} = \arg_{\theta}\max \log \tilde{p}(\theta|{\cal Y})$ and ${\bf H}^{*}_{\theta}$. Finally we do a grid search from the 
mode in all directions until $\log\tilde{p}({\theta}^{*}|{\cal Y})-\log\tilde{p}(\theta|{\cal Y})>\varphi$, for a given threshold 
$\varphi$.

\section{Proposed Approach}
\label{section: Proposed Approach}

Our proposed algorithm estimates the underlying densities for the number of neighbours of probabilistic kNN classification by 
using Eq. (\ref{eq: marginal posterior of theta in INLA}). To distinguish it from other model selection approaches, we term this
approach \emph{KOREA}, which is an acronym for "{\bf K}-{\bf OR}der {\bf E}stimation {\bf A}lgorithm" in a Bayesian framework.

\subsection{Obtaining the optimal number of neighbours $K^{*}$}
Let ${\cal Y}$ denote a set of observations and let ${\bf f}_{K}$ be a set of the model parameters given a model order $K$. The first step of \emph{KOREA} is to estimate the optimal number of neighbours, $K^{*}$:
\begin{eqnarray}
K^{*} = \arg_{K}\max p(K|{\cal Y}). 
\end{eqnarray}
According to Eq. (\ref{eq: marginal posterior of theta in INLA}), we can obtain an approximated marginal posterior distribution by
\begin{equation}
\tilde{p}(K|{\cal Y}) \propto \left.\frac{p({\cal Y}, {\bf f}_{K}, K)}{p_{F}({\bf f}_{K}|{\cal Y}, K)}\right|_{{\bf f}_{K}(K)={\bf f}_{K}^{*}(K)}.
\label{eq: the general format of KOREA}
\end{equation}
This equation has the property that $K$ is an integer variable while $\theta$ of Eq. (\ref{eq: marginal posterior of theta in INLA}) 
is in general a vector of continuous variables. By ignoring the difference, we can still use the Quasi-Newton method to efficiently 
obtain optimal $K^{*}$. Alternatively, we can also calculate some potential candidates between $1$ and $K_{\max}$ if $K_{\max}$ is 
not too large. Otherwise, we may still use the \emph{Quasi-Newton} style algorithm with a rounding operator which transforms a 
real value to an integer for $K$.

\subsection{Bayesian Model Selection for PKNN classification}

In general, one of the most significant problems in classification is to infer the joint posterior distribution of $L$ different 
hidden classes for $L$ different observations such that 
${\bf z}_{1:L}^{'*} = \arg_{{\bf z}_{1:L}^{'}}\max p({\bf z}_{1:L}^{'}|{\bf y}, {\bf z}, {\bf y}_{1:L}^{'})$. However, jointly 
inferring the hidden variables is not straightforward therefore we make the assumption that the hidden class of the $i$-th 
observation $z_{i}^{'}$ is independent to one of the $j$-th observation given the $i$-th observation ${\bf y}_{i}^{'}$ where 
$i\neq j$ and then we have the following simpler form (similar to Naive Bayes):
\begin{equation}
p({\bf z}_{1:L}^{'}|{\bf y}, {\bf z}, {\bf y}_{1:L}^{'})  = \prod_{i=1}^{L}p(z_{i}^{'}|{\bf y}, {\bf z}, {\bf y}_{i}^{'})
\label{eq: full marginal posterior}
\end{equation}
where $p(z_{i}^{'}|{\bf y}, {\bf z}, {\bf y}_{i}^{'})$ is estimated by Eq. (\ref{eq: prediction of PKNN}).

\subsubsection{PKNN via KOREA}
In the probabilistic kNN model (PKNN), let us define the new dataset with $L$ data by ${\bf y}_{1:L}^{'}$, which is not labeled yet. 
The unknown labels are denoted by ${\bf z}_{1:L}^{'}$. Here we use ${\bf y}_{i}^{'}$ and $z_{i}^{'}$ for the $i$th new observation 
and its hidden label. That is, we have a hidden variable ${\bf f}_{K} = z_{i}^{'}$ of interest given 
${\bf z}={\bf z}_{1:N}$, ${\bf y} = {\bf y}_{1:N}$ and ${\bf y}_{i}^{'}$ such that 
${\cal Y} = ({\bf z}, {\bf y}, {\bf y}_{i}^{'})$. The target posterior is obtained in a similar form to 
Eq. (\ref{eq: the general format of KOREA}) as

\begin{eqnarray}
p(z_{i}^{'}|{\bf y}, {\bf z}, {\bf y}_{i}^{'}) &=& p(z_{i}^{'}|{\cal Y})= \int_{K, \beta}p(z_{i}^{'}, \beta, K|{\cal Y})d\beta dK\cr
&=& \int p(z_{i}^{'}| \beta, {\cal Y}, K)p(\beta|{\cal Y}, K)p(K|{\cal Y})d\beta dK\cr
&\approx& \sum_{\beta^{(m)}}\sum_{j=1}^{K_{\max}}\left[p(z_{i}^{'}| \beta^{(i)}, {\cal Y}, K=j) p(\beta^{(m)}|{\cal Y}, K=j)\right.\cr
&&\times \left. p(K=j|{\cal Y}) \Delta_{\beta^{(m)}}\right]\cr
&\approx& \sum_{\beta^{(m)}}\sum_{j=1}^{K_{\max}}\left[p(z_{i}^{'}| \beta^{(i)}, {\cal Y}, K=j) \tilde{p}(\beta^{(m)}|{\cal Y}, K=j)\right.\cr
&& \times \left.\tilde{p}(K=j|{\cal Y}) \Delta_{\beta^{(m)}}\right]\cr
&=& \sum_{\beta^{(m)}}\sum_{j=1}^{K_{\max}}\lambda_{j}^{(m)}p(z_{i}^{'}| \beta^{(m)}, {\cal Y}, K=j)
\label{eq: prediction of PKNN}
\end{eqnarray}
where
\begin{equation}
\lambda_{j}^{(m)} = \frac{\tilde{p}(\beta^{(m)}|{\cal Y}, K=j)\tilde{p}(K=j|{\cal Y}) \Delta_{\beta^{(m)}}}
{
\sum_{\beta^{(a)}}\sum_{b=1}^{K_{\max}}\tilde{p}(\beta^{(a)}|{\cal Y}, K=b)\tilde{p}(K=b|{\cal Y}) \Delta_{\beta^{(a)}}
}.
\label{eq: weight for the PKNN}
\end{equation}

Now we need to know three distributions in the above equation. 
\begin{enumerate}
\item $p(z_{i}^{'}|\beta^{(m)}, {\cal Y}, K=i)$: conditional likelihood
\item $\tilde{p}(\beta^{(m)}|{\cal Y}, K=i)$: posterior of $\beta$
\item $p(K|{\cal Y})$: posterior of $K$
\end{enumerate}
The first equation among the three above is the conditional distribution and it is defined by
\begin{equation}
p(z_{i}^{'}| \beta^{(m)}, {\cal Y}, K=j) = \frac{p({\bf z}, z_{i}^{'}| \beta^{(m)}, {\cal Y}, K=j)}{\sum_{c\in{\bf C}}p({\bf z}, z_{i}^{'}=c| \beta^{(m)}, {\cal Y}, K=j)}.
\end{equation}
This is a likelihood function given the neighbouring structure. That is, $p({\bf z}, z^{'}|$ $\beta^{(m)}, {\cal Y}, K=j)$ explains the fitness between the assumed/given labels $({\bf z}, z_{i}^{'})$ and the given full data (${\bf y}, {\bf y}_{i}^{'}$)

Another equation is $\tilde{p}(\beta^{(m)}|{\cal Y}, K=j)$ but we defer the estimation of this distribution since it can be 
automatically estimated when we estimate the last distribution $p(K|{\cal Y})$. Therefore, we infer the last equation first. The 
last equation is the marginal posterior of $K$ and using a similar approach to INLA it is defined by
\begin{eqnarray}
\tilde{p}(K|{\cal Y}) &\propto& \left.
\frac{
p({\bf z}, \beta, K|{\bf y}, {\bf y}_{i}^{'})
}{
p_{G}(\beta|{\bf z}, {\bf y}, {\bf y}_{i}^{'}, K)
}\right|_{\beta = \beta^{*}(K)}
= \left.
\frac{
\sum_{c\in{\bf C}}p(z_{i}^{'}, {\bf z}, \beta, K|{\bf y}, {\bf y}_{i}^{'})
}{
p_{G}(\beta|{\bf z}, {\bf y}, {\bf y}_{i}^{'}, K)
}\right|_{\beta = \beta^{*}(K)}\cr
&=& \left.
\frac{
p(\beta)p(K)\sum_{c\in{\bf C}}p(z_{i}^{'}=c, {\bf z}| \beta, K, {\bf y}, {\bf y}_{i}^{'})
}{
p_{G}(\beta|{\bf z}, {\bf y}, {\bf y}_{i}^{'}, K)
}\right|_{\beta = \beta^{*}(K)}.
\end{eqnarray}
As we can see the denominator is the approximation of the second distribution of interest so we can reuse it i.e. 
$\tilde{p}(\beta|{\cal Y}, K)=p_{G}(\beta|{\bf z}, {\bf y}, {\bf y}_{i}^{'}, K)$ which is a Gaussian approximation of 
$p(\beta|{\cal Y}, K)\propto p({\bf z}|{\bf y}, {\bf y}_{i}^{'}, K)p(\beta)=\sum_{c\in{\bf C}}p(z_{i}^{'}=c, 
{\bf z}|{\bf y}, {\bf y}_{i}^{'}, K)p(\beta)$.

We also easily obtain the marginal posterior of $\beta$ which is $p(\beta|{\cal Y})$. Since the marginal posterior is approximated
by $p(\beta|{\cal Y})\approx \tilde{p}(\beta|{\cal Y})=\sum_{j=1}^{K_{\max}}\tilde{p}(\beta|{\cal Y}, K=j)\tilde{p}(K=j|{\cal Y})$, 
we can simply reconstruct the distribution by reusing the previously estimated distributions. When we have 
$\mu_{\beta}^{(j)}={\bf E}(\beta|{\cal Y}, K=j)$ and $\sigma_{\beta}^{(j)2}={\bf V}(\beta|{\cal Y}, K=j)$, then we have
\begin{equation}
\mu_{\beta} = \sum_{j=1}^{K_{\max}}\tilde{\alpha}_{j} \mu_{\beta}^{(j)} \textrm{ and }
\sigma_{\beta}^{2} = \sum_{j=1}^{K_{\max}}\tilde{\alpha}_{j} \left[\sigma_{\beta}^{(j)2}+\left\{\mu_{\beta} - \mu_{\beta}^{(j)}  \right\}^{2}\right].
\label{eq: marginal posterior of beta}
\end{equation}

Finally, we can obtain the target distribution of interest $p(z_{i}^{'}|{\bf y}, {\bf z}, {\bf y}_{i}^{'})$ with three 
distributions. Since we can now estimate the target distribution as a mixture distribution, we can also obtain the expectation 
and variance as follows:
\begin{eqnarray}
{\bf E}(z_{i}^{'}|{\cal Y}) &=& \sum_{\beta^{(m)}}\sum_{j=1}^{K_{\max}}\lambda_{j}^{(m)}\mu_{m, j}^{(i)}\cr
{\bf V}(z_{i}^{'}|{\cal Y}) &=& \sum_{\beta^{(m)}}\sum_{j=1}^{K_{\max}}\lambda_{j}^{(m)}\left[
\Sigma_{m,j}^{(i)} + \left\{{\bf E}(z_{i}^{'})-\mu_{m, j}^{(i)}\right\}^{2}
\right]
\label{eq: mean and variance of predicted labels}
\end{eqnarray}
where $\mu_{m, j}^{(i)} = {\bf E}(z_{i}^{'}|{\cal Y}, \beta^{(m)}, K=j)$ and 
$\Sigma_{m,j}^{(i)} = {\bf V}(z_{i}^{'}|{\cal Y}, \beta^{(m)}, K=j)$. Here 
$p(\beta)=\mathcal{G}(\beta; a_{\beta}, b_{\beta})$ and $\mathcal{IG}(\cdot; a, b)$ represents inverse Gamma distribution 
with hyper-parameters $a$ and $b$. In this paper, we set $a=2$ and $b=10$ yielding an almost flat prior.

\begin{algorithm}[h!]
\caption{PKNN classifier via \emph{KOREA}}
\label{algorithm: Generic framework of PKNN classifier via KOREA}
\begin{algorithmic}[1]
\REQUIRE Given $N$ observations, $({\bf y}, {\bf z})=({\bf y}_{1:N}, {\bf z}_{1:N})$, a new testing set with $L$ observations, ${\bf y}^{'}={\bf y}_{1:L}^{'}$ and a set of classes ${\bf C}$

\FOR{$i=1$ to $L$}
	\STATE Obtain a new observation ${\bf y}_{i}^{'}$ and set ${\cal Y}=({\bf y}, {\bf z}, {\bf y}_{i}^{'}, \hat{\beta})$.
	\newline
	{\bf - Calculate $\tilde{p}(K|{\cal Y}, \hat{\beta})$.}
	\FOR{$j=1$ to $K_{\max}$}
		\STATE Calculate the approximate conditional posterior $\tilde{p}(\beta|{\cal Y}, K=j)= p_{G}(\beta|{\cal Y}, K=j)$ by using Gaussian approximation of $p(\beta|{\cal Y}, K=j)\propto p({\bf z}|{\bf y}, {\bf y}_{i}^{'}, \beta, K=j)p(\beta)$.
		\STATE Obtain $\mu_{\beta}^{(j)}={\bf E}(\beta|{\cal Y}, K=j)$ and $\sigma_{\beta}^{(j)2}= {\bf V}(\beta|{\cal Y}, K=j)$.
		\STATE Calculate an unnormalized posterior for $K=j$,
		$
		\alpha_{j} = \tilde{p}(K=j|{\cal Y})
				 \propto \frac{p(\mu_{\beta}^{(j)})p(K=j)\sum_{c\in{\bf C}}p(z_{i}^{'}=c, {\bf z}|\mu_{\beta}^{(j)}, K=j, {\bf y}, {\bf y}_{i}^{'})}{p_{G}(\mu_{\beta}^{(j)}|{\cal Y}, K=j)}.
		$
	\ENDFOR
	\STATE Normalize the model order weights by $\tilde{\alpha}_{s}=\frac{\alpha_{s}}{\sum_{j=1}^{K_{\max}}\alpha_{j}}$ for all $s\in \{1, 2, \cdots, K_{\max}\}$. 
	\STATE Calculate the mean $\mu_{\beta}$ and variance $\sigma_{\beta}^{2}$ of marginal posterior of $\beta$ from Eq. (\ref{eq: marginal posterior of beta}).
	\STATE ${\bf S}_{\beta}=\{\beta| 0<\beta=\mu_{\beta}\pm i\sigma_{\beta}< \beta_{\max} \textrm{ for }i=1, 2, \cdots \}$.
	\STATE Calculate an unnormalize weight $\lambda_{j}^{(m)}=\tilde{p}(\beta^{(m)}|{\cal Y}, K=j)\alpha_{j}$ for $j=1,2, \cdots, K$ and $m=1, 2, \cdots, |S_{\beta}|$.
	\STATE Obtain $\tilde{\lambda}^{(m)}_{j}= \frac{\lambda_{j}^{(m)}}{\sum_{n=1}^{|{\bf S}_{\beta}|}\sum_{k=1}^{K_{\max}}\lambda_{k}^{(n)}}$ for all $j\in\{1,2,\cdots, K_{\max}\}$ and all $m\in\{1, 2, \cdots, |{\bf S}_{\beta}|\}$  from Eq. (\ref{eq: weight for the PKNN}).
	\newline
	{\bf - Calculate the solution of $p(z_{i}^{'}|{\cal Y})$.}
	\FOR{$m=1$ to $|{\bf S}_{\beta}|$}
		\FOR{$j=1$ to $K_{\max}$}
			\STATE {\bf for}{$c\in {\bf C}$} {\bf Get $\tau_{j,c}=p(z_{i}^{'}=c, {\bf z}|{\bf y}, {\bf y}_{i}^{'}, K=j, \beta^{(m)})$.}
			\STATE {\bf for}{$c\in {\bf C}$} {\bf Get $\tau_{j,c}^{(m)}=\frac{\tau_{j, c}}{\sum_{l\in{\bf C}} \tau_{j, l} }$.}
		\ENDFOR
	\ENDFOR
	\STATE Calculate $p(z_{i}^{'}=c|{\cal Y})= \sum_{m=1}^{|{\bf S}_{\beta}|}\sum_{k=1}^{K_{\max}} \tau_{k,c}^{(m)}\tilde{\lambda}_{k}^{(m)}$ for all $c\in{\bf C}$.
	\STATE Calculate the expectation and variance of $z_{i}$ from Eq. (\ref{eq: mean and variance of predicted labels}).
\ENDFOR
\end{algorithmic}
\end{algorithm}

\subsection{Additional Neighbouring Rules}

\subsubsection{A Boltzmann modelling with equal weights}
In the conventional Boltzmann modelling for the neighbouring 
structure, the interaction rate $\beta$ is divided by a fixed $K$ as shown in 
Eq. (\ref{eq: approximated symmetirc likelihood of PKNN}). This results in each neighbour having its own different weight. 
Therefore, we need to apply an equal weight to the neighbours by varying $K$ for the different neighbouring structure. In order 
to build this strategy, we adopt three sequential approaches: (i) obtain a neighbour structure in the same way as conventional 
Boltzmann modelling; (ii) modify the structure by transforming from a directed graph to an undirected graph. If $j\in ne(i)$ but 
$i\notin ne(j)$ then we add $i$ into $ne(j)$ for $i\neq j$; and (iii) apply the pseudo likelihood for the likelihood. In this paper, we name this modelling as 
\emph{Boltzmann}$^{(2)}$ modelling.

\section{Simulation Results}
\label{section: Simulation Results}

The performance of our algorithm is tested with a collection of benchmark datasets. All of the datasets (test and training) used in this paper can be found at \url{http://mathsci.ucd.ie/~nial/dnn/}. The six well-known benchmark datasets are presented in 
Table \ref{table: description of benchmark}. 
\begin{table}[h!]
\caption{Benchmark datasets: $C$ (the number classes), $d$ (the dimension of the data), $N_{total}$ (the total number of data)}
\label{table: description of benchmark}
\centering
\begin{tabular}{c|ccc}
\hline\hline
Name of data & $C$ & d & $N_{total} $ \cr
\hline
Crabs 		& 4 		& 5		& 200 		\cr
Fglass		& 4 		& 9		& 214 		\cr
Meat 		& 5 		& 1050	& 231 		\cr
Oliveoil		& 3 		& 1050	&	65 		\cr
Wine			& 3		& 13		&	178		\cr
Iris			& 3		& 4		& 	150		\cr
\hline\hline
\end{tabular}
\end{table}
We test the performance by using $4$-fold cross validation for a fair comparison with all approaches although our proposed approach does not not require it due to the Bayesian nature of it.
\begin{figure}[h!]
\centering
\begin{tabular}{lll}
\includegraphics[scale=0.2]{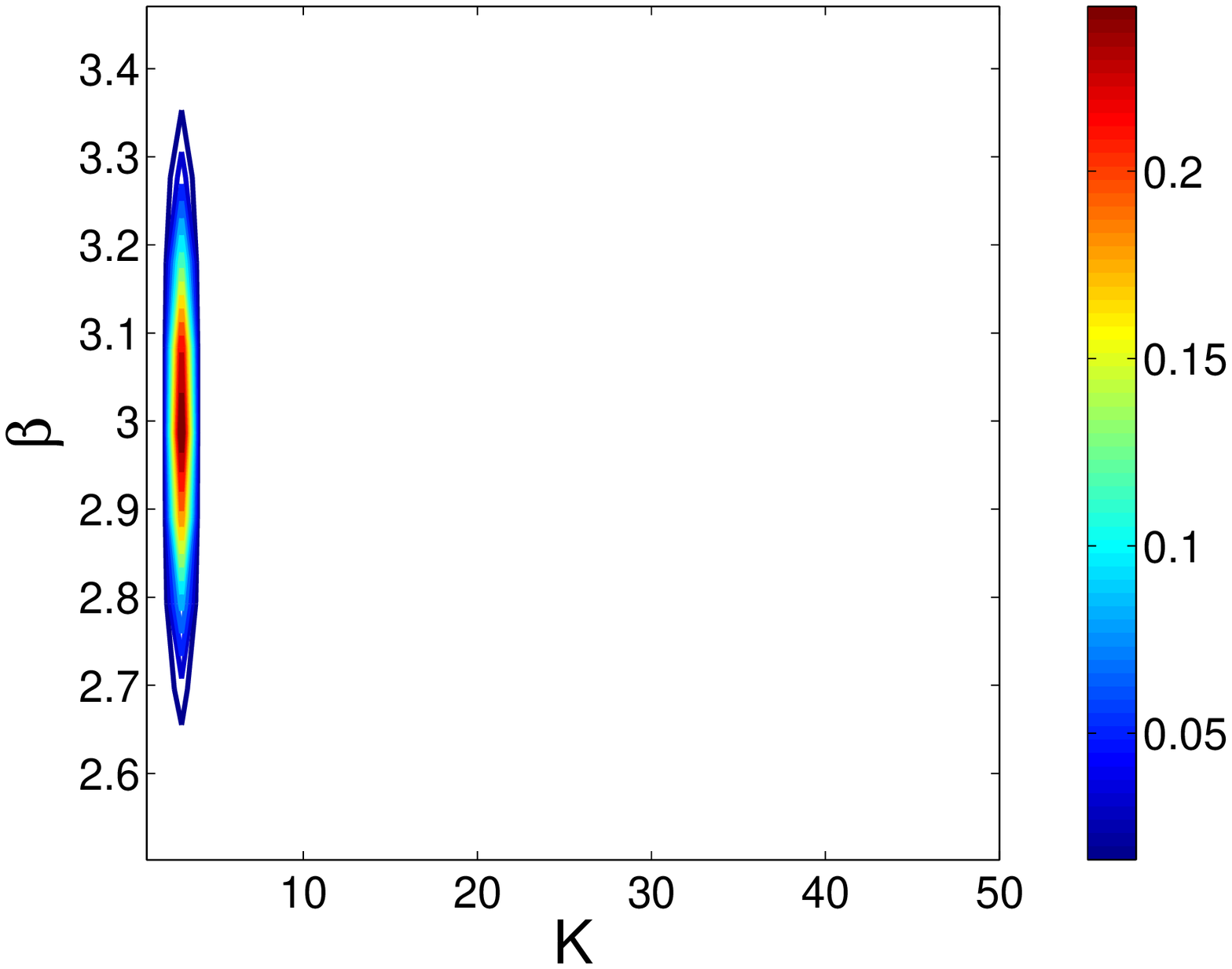}&
\includegraphics[scale=0.2]{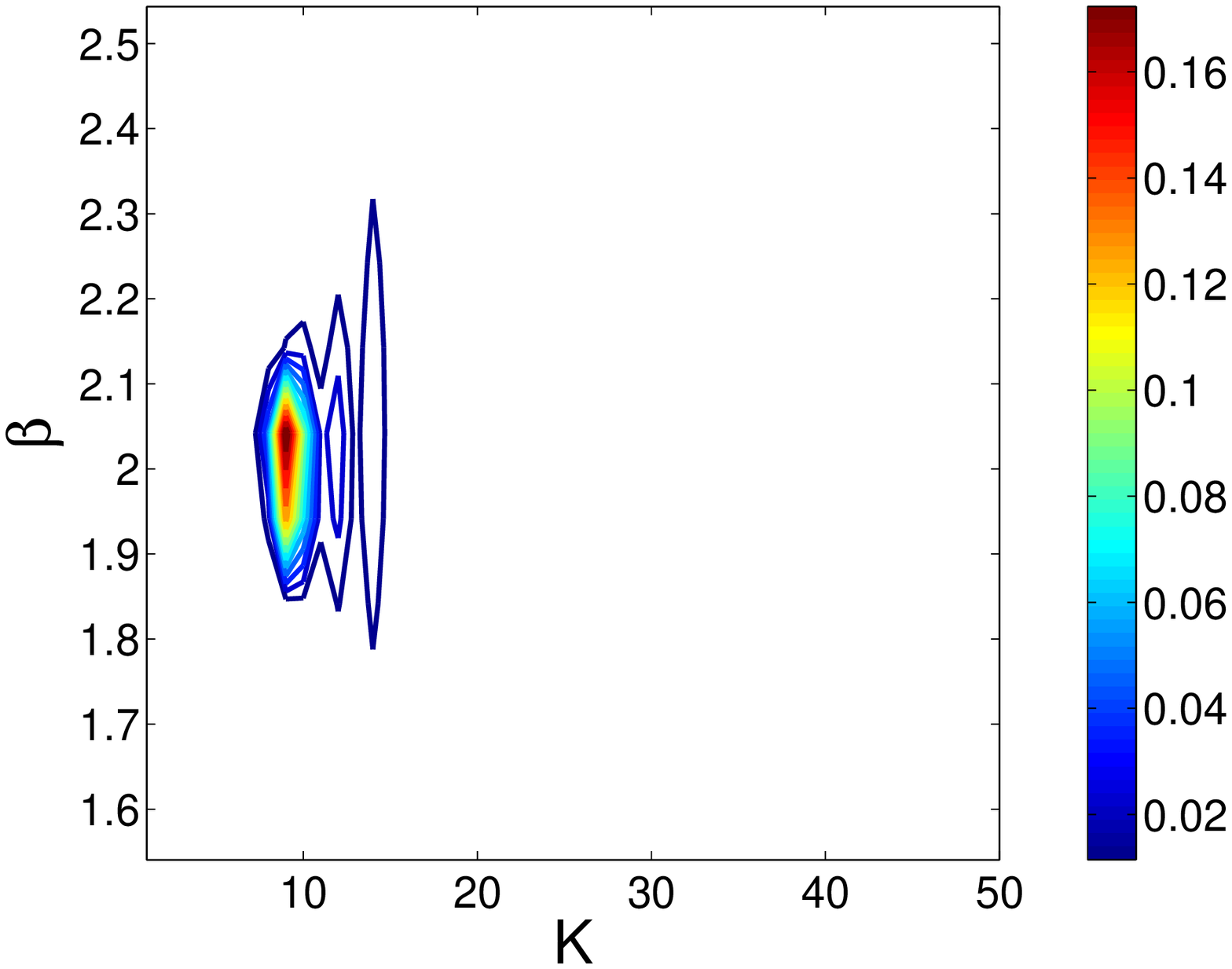}&
\includegraphics[scale=0.2]{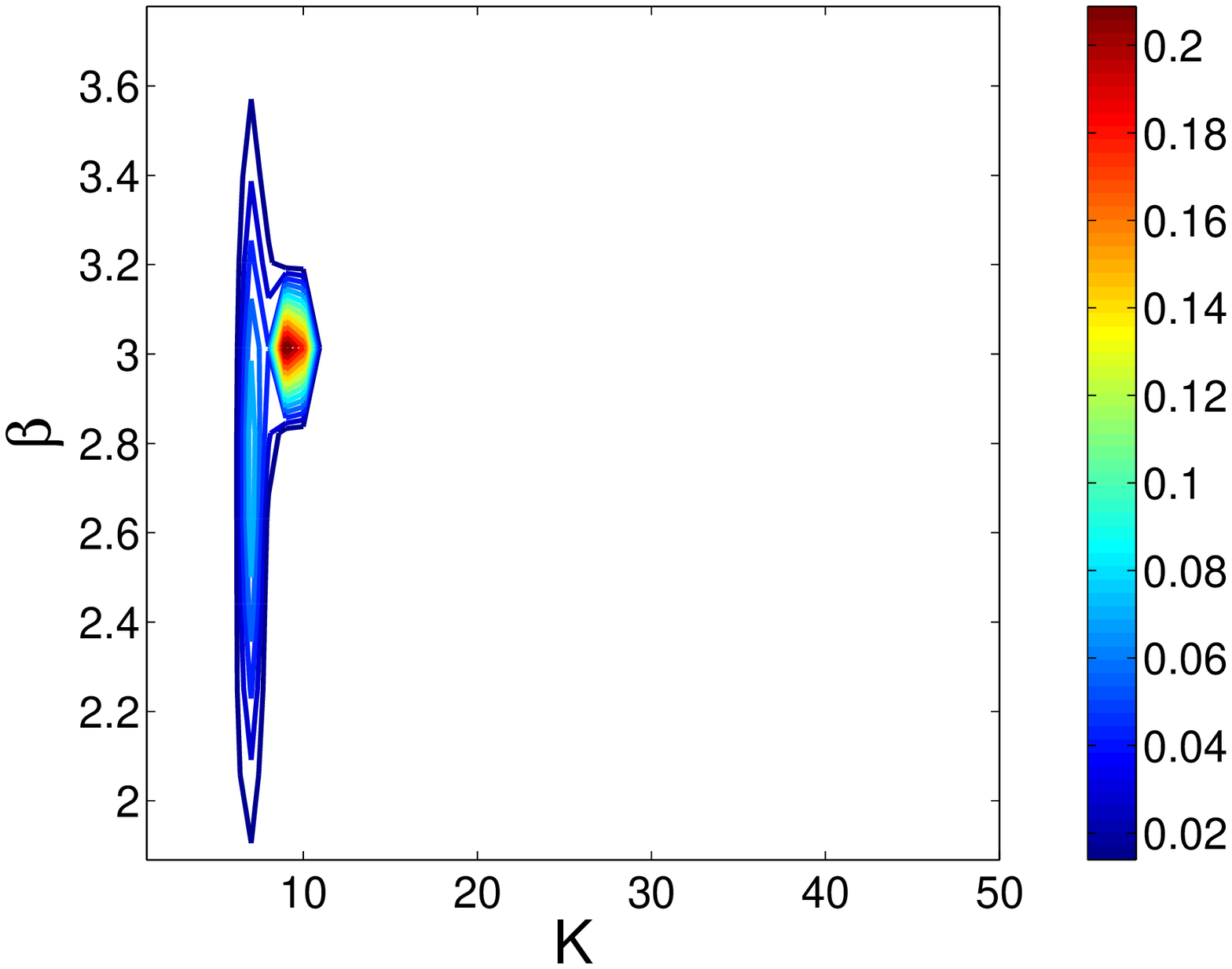}\cr
\includegraphics[scale=0.2]{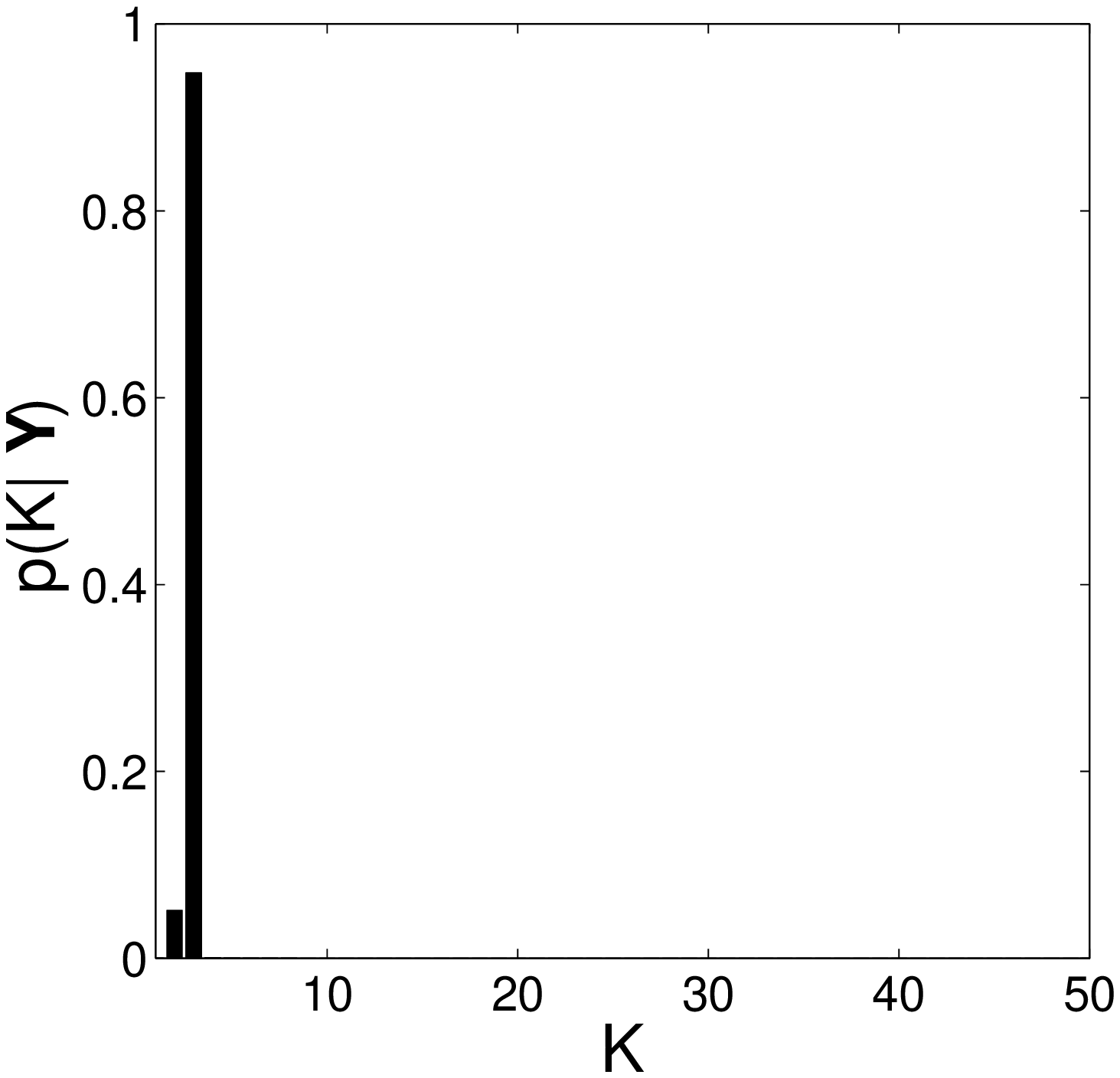}&
\includegraphics[scale=0.2]{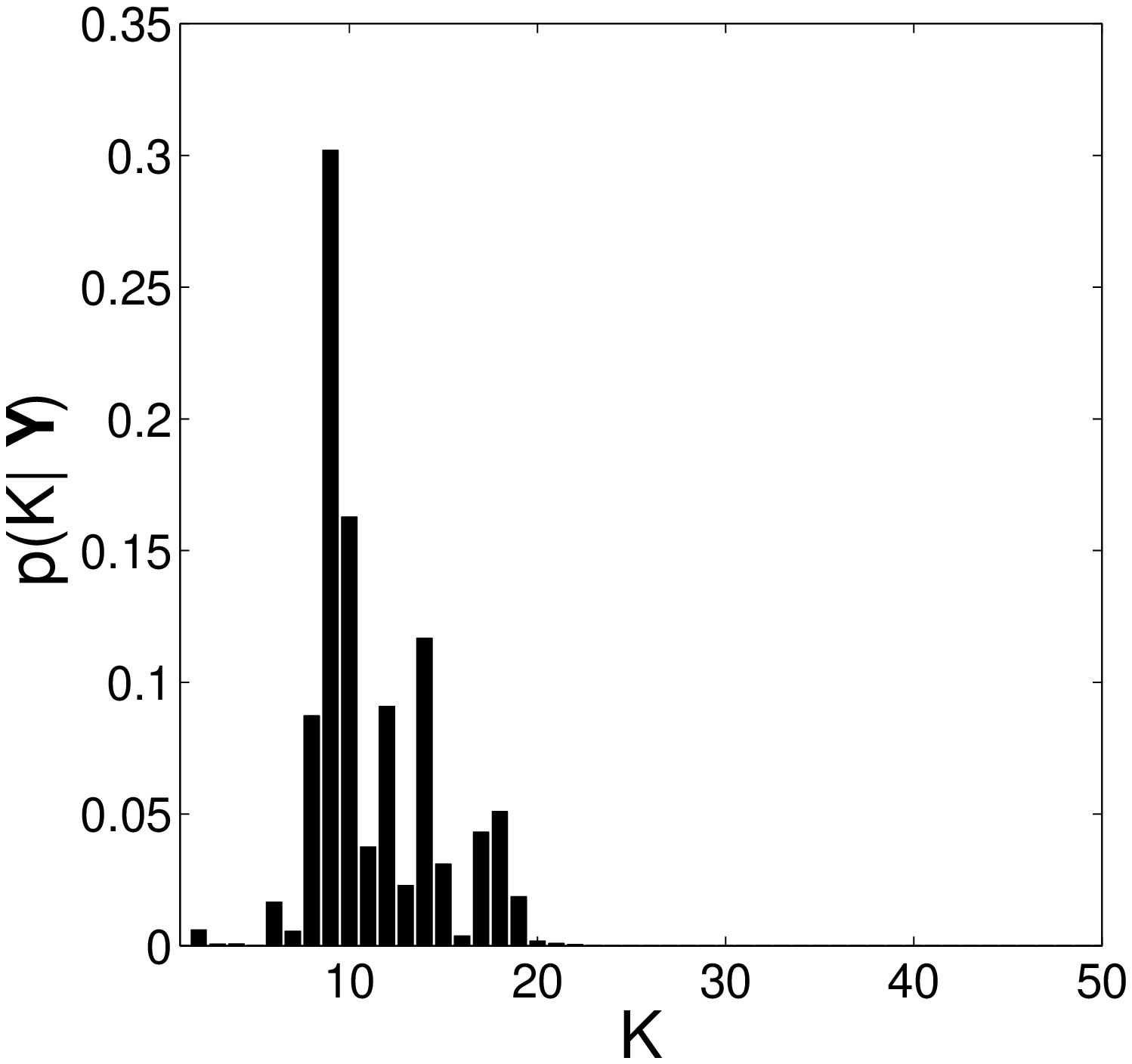}&
\includegraphics[scale=0.2]{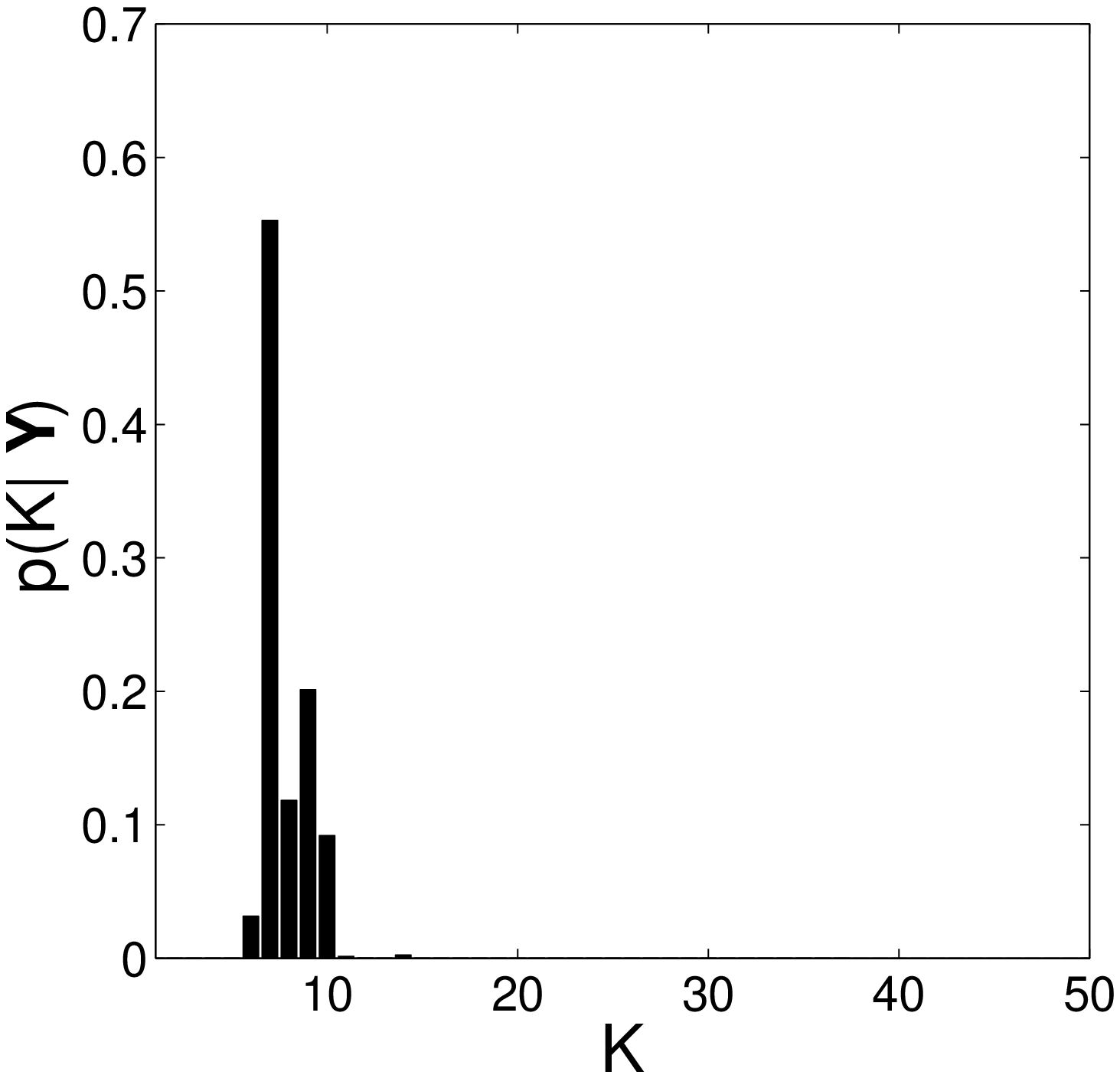}\cr
(a) Crabs & (b) Fglass & (c) Meat\cr
\hline\hline
\includegraphics[scale=0.2]{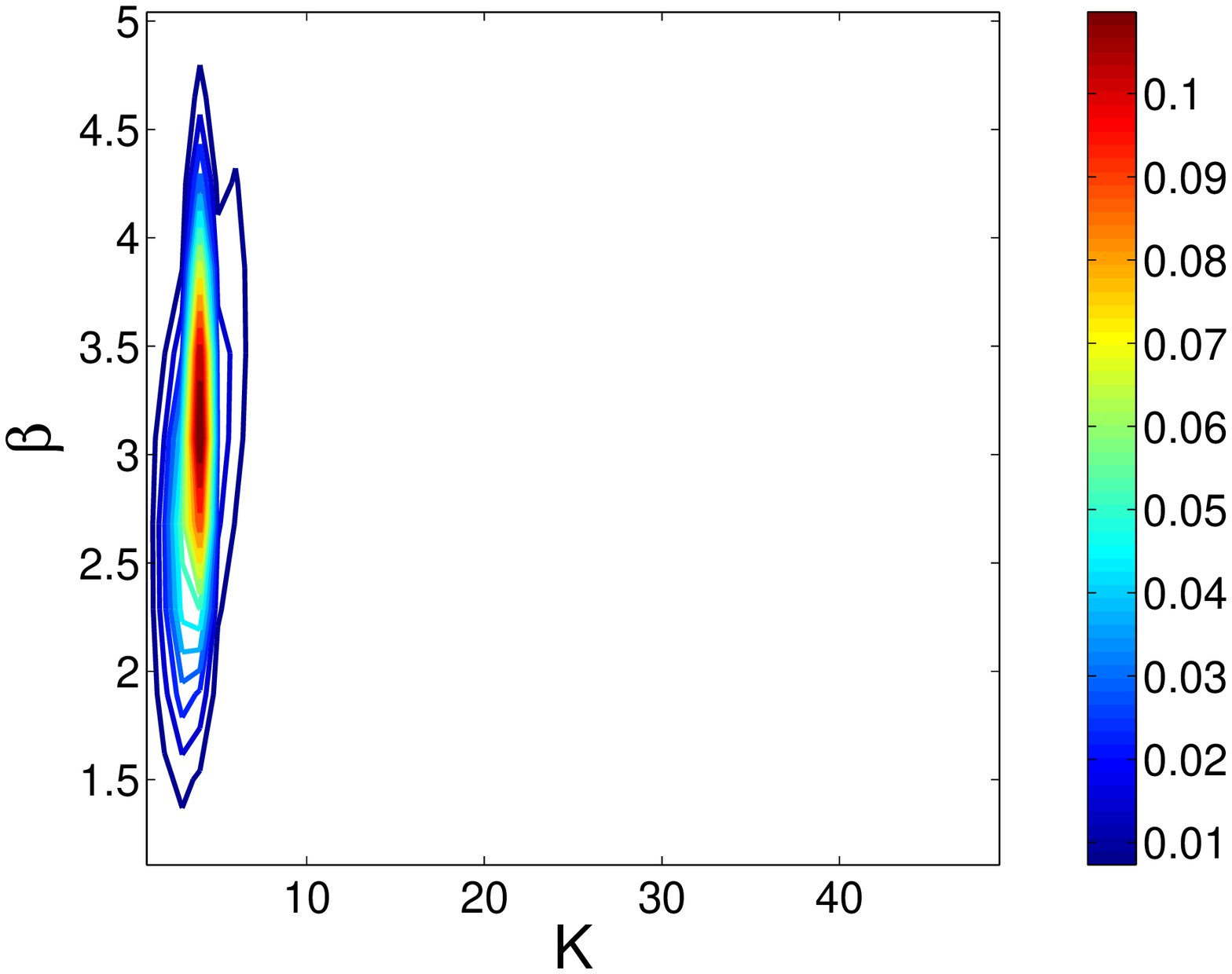}&
\includegraphics[scale=0.2]{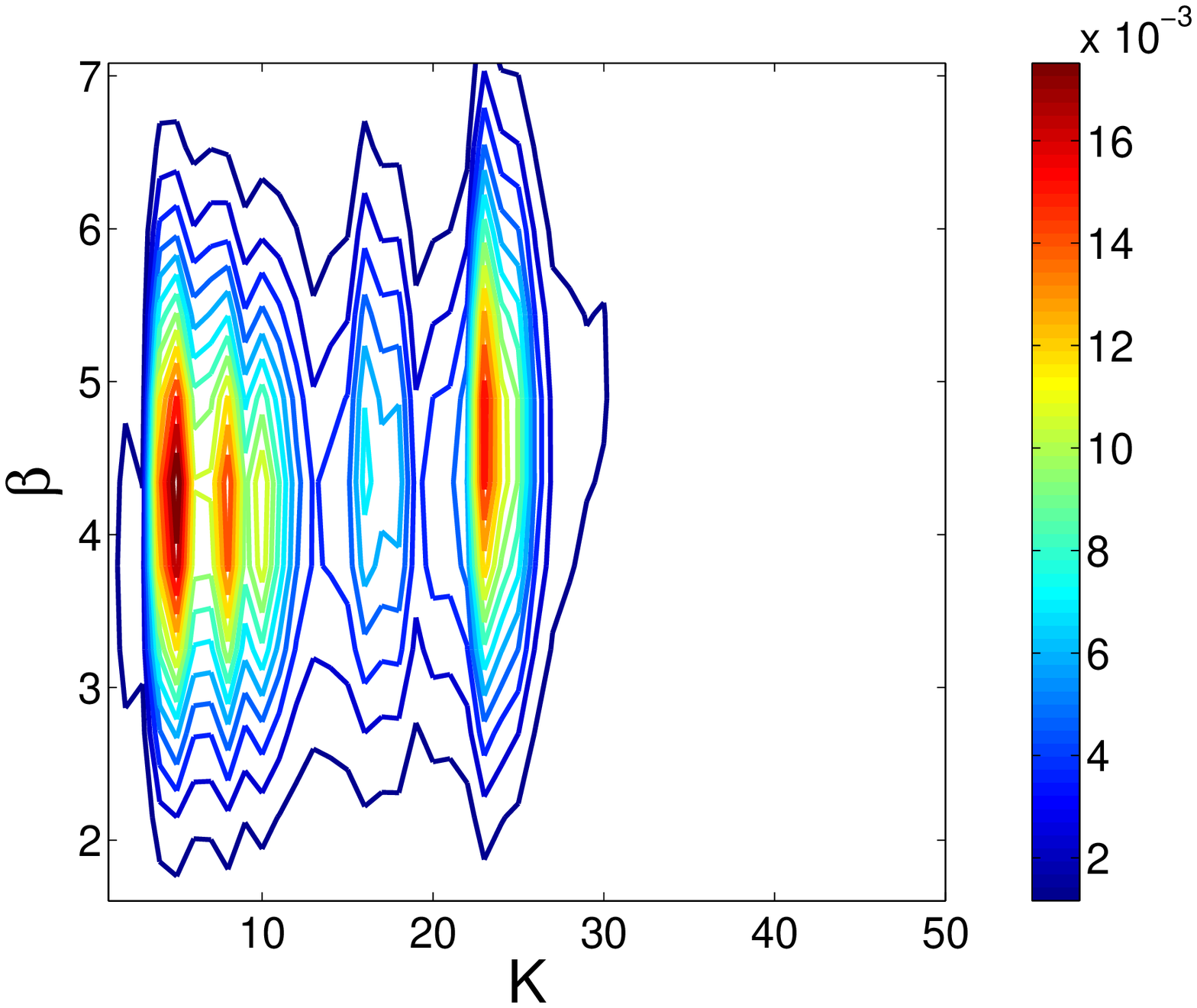}&
\includegraphics[scale=0.2]{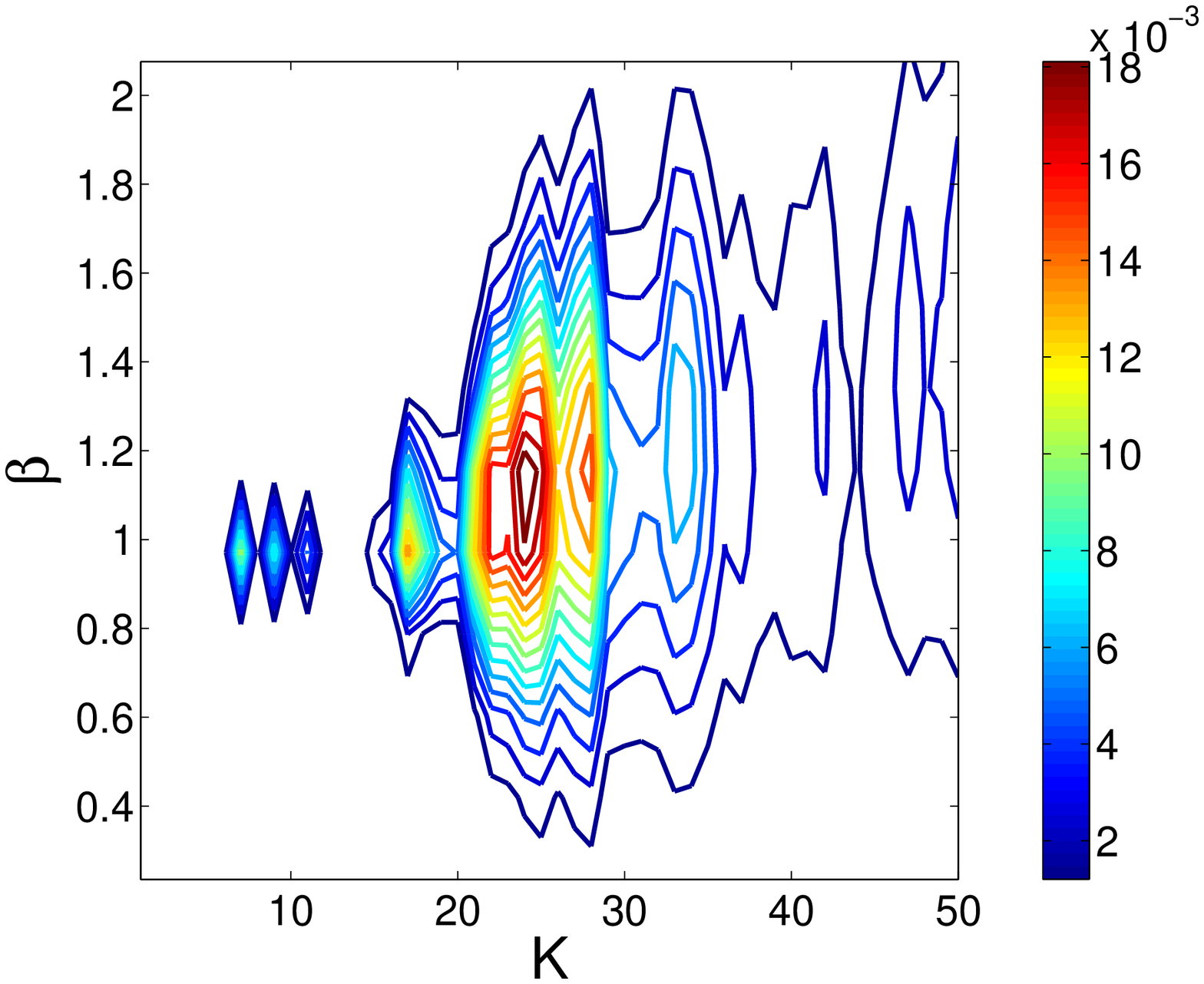}\cr
\includegraphics[scale=0.2]{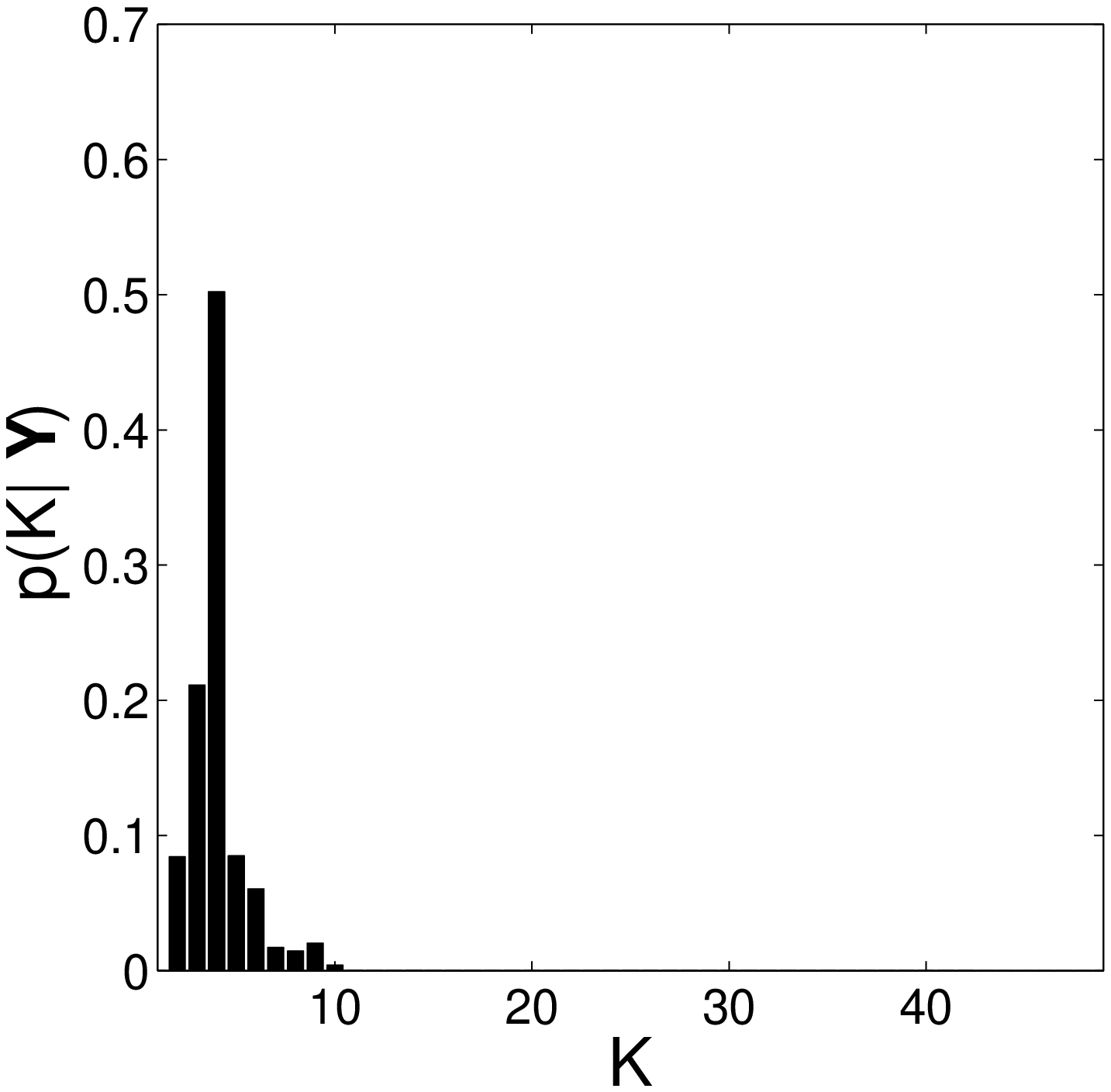}&
\includegraphics[scale=0.2]{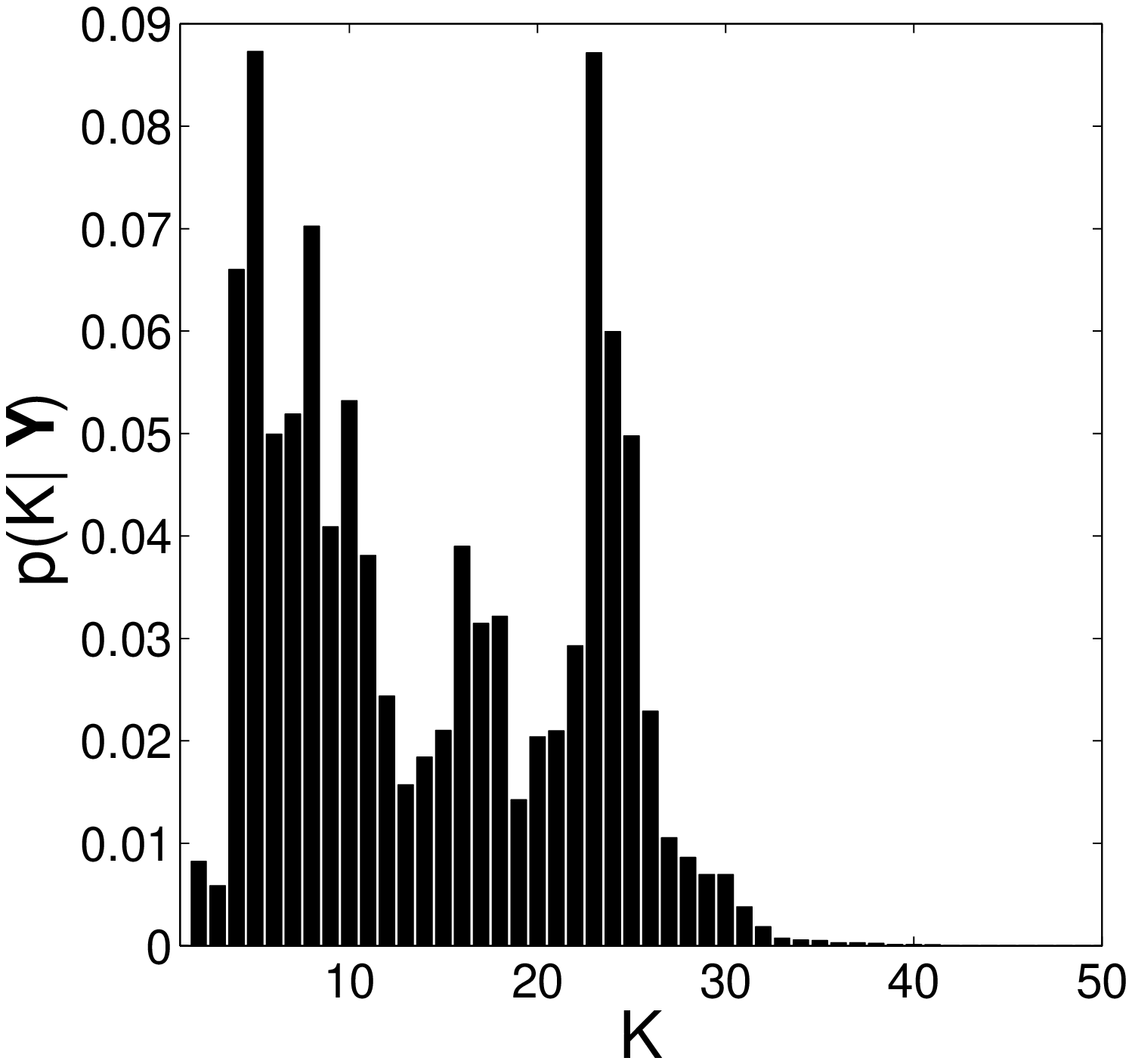}&
\includegraphics[scale=0.2]{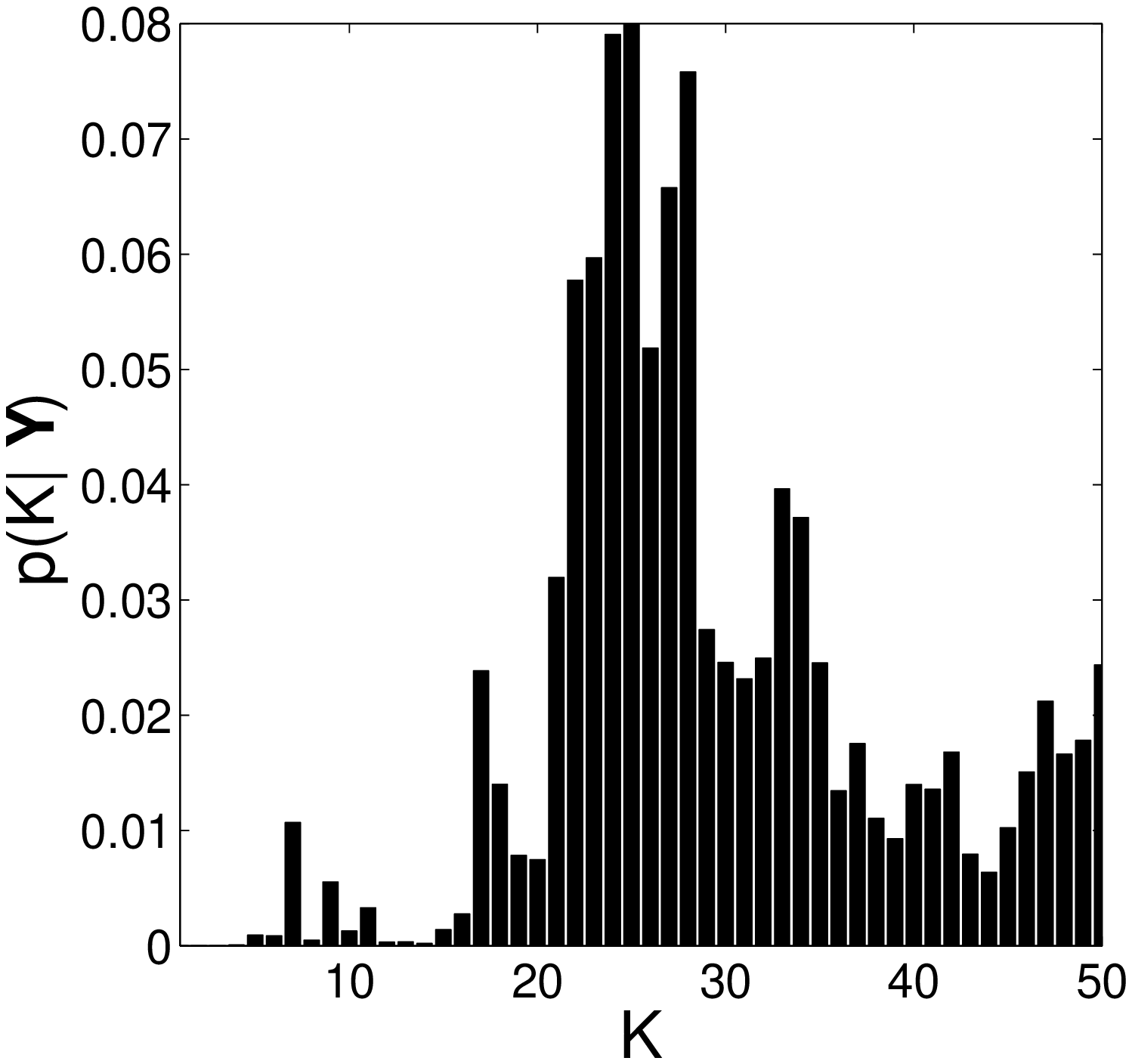}\cr
(d) Oliveoil & (e) wine & (f) Iris\cr
\end{tabular}
\caption{Posterior distribution $p(K, \beta|{\cal Y})$ [top] and its marginalized posterior density $p(K|{\cal Y})$ [bottom] via KOREA}
\label{fig: Posterior distribution p(K, beta|Y)}
\end{figure}

Figure \ref{fig: Posterior distribution p(K, beta|Y)} demonstrates reconstructed densities of a testing datum. While top subgraphs show the 2 dimensional densities $p(\beta, K|{\cal Y})$, bottom sub-figures represent the 1 dimensional densities $p(K|{\cal Y})$ 
for all datasets. The graphs illustrate that the distribution is not unimodal but a complex multi-modal distribution. This also suggests that selecting an appropriate number of neighbours for PKNN is critical to obtain high accuracy.

\begin{figure}[h!]
\centering
\begin{tabular}{c}
\includegraphics[width=5in, height=3in]{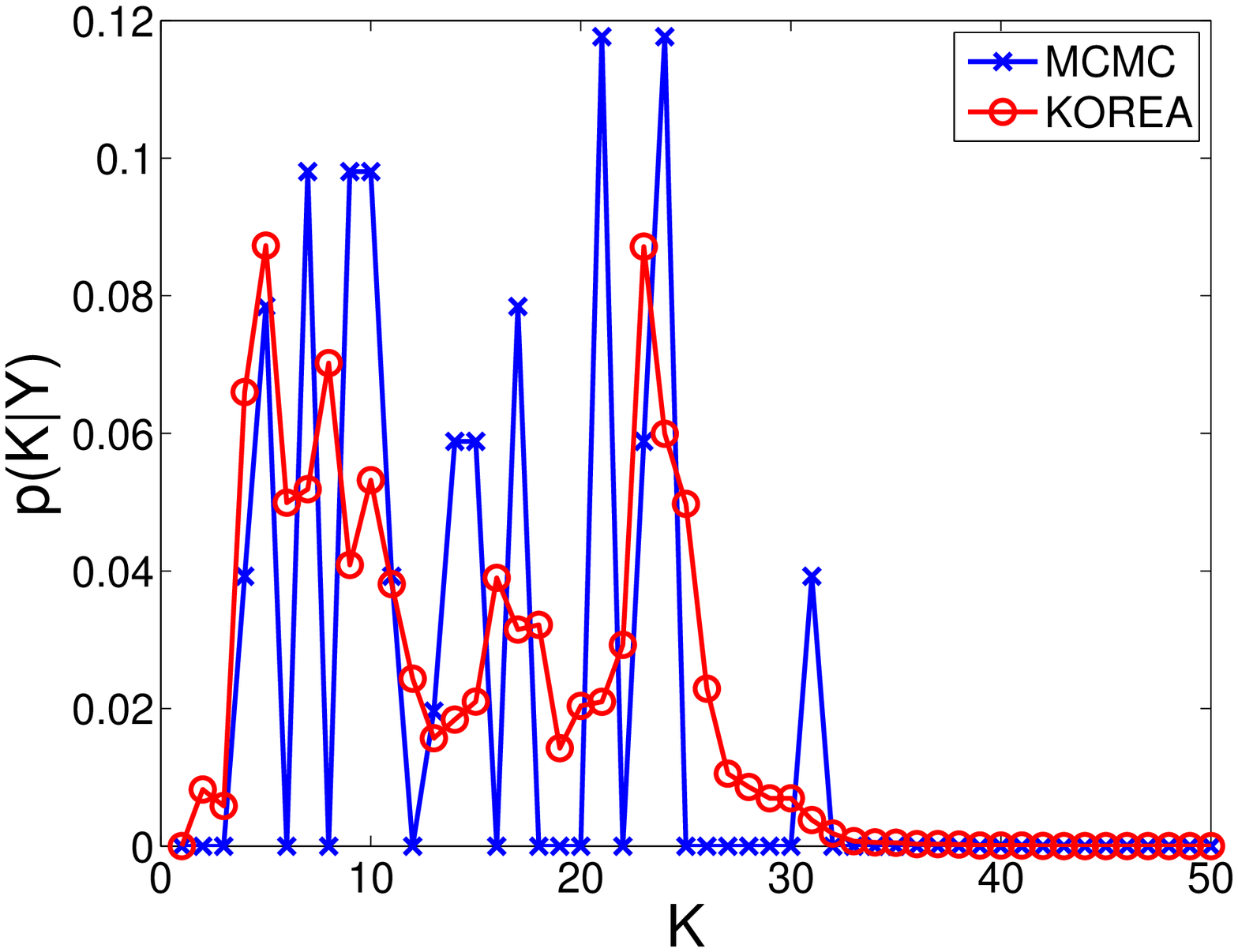}\cr
(a) $N_{MCMC}=100$\cr
\includegraphics[width=5in, height=3in]{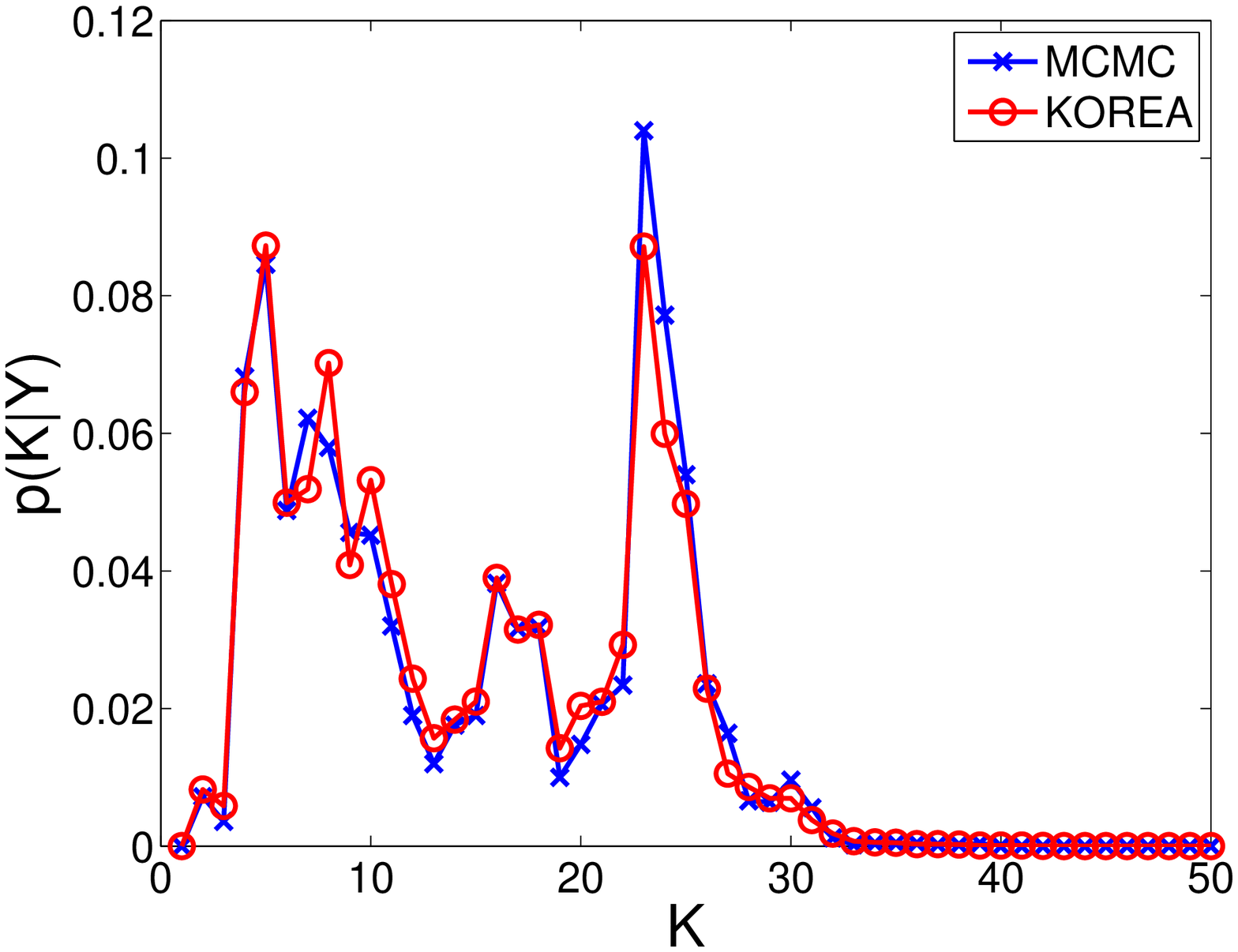}\cr
(b) $N_{MCMC}=10000$
\end{tabular}
\caption{Comparison between MCMC and KOREA for wine dataset, $N_{MCMC}$ denotes the number of MCMC iterations.}
\label{fig: Comparison between MCMC and KOREA for wine dataset}
\end{figure}

\begin{figure}[h!]
\centering
\fontsize{8}{12}\selectfont
\begin{tabular}{ccccc}
		& RMSE 	& KLD 	& PSNR 	& SSIM \cr
Crabs	& 
\includegraphics[scale=0.13]{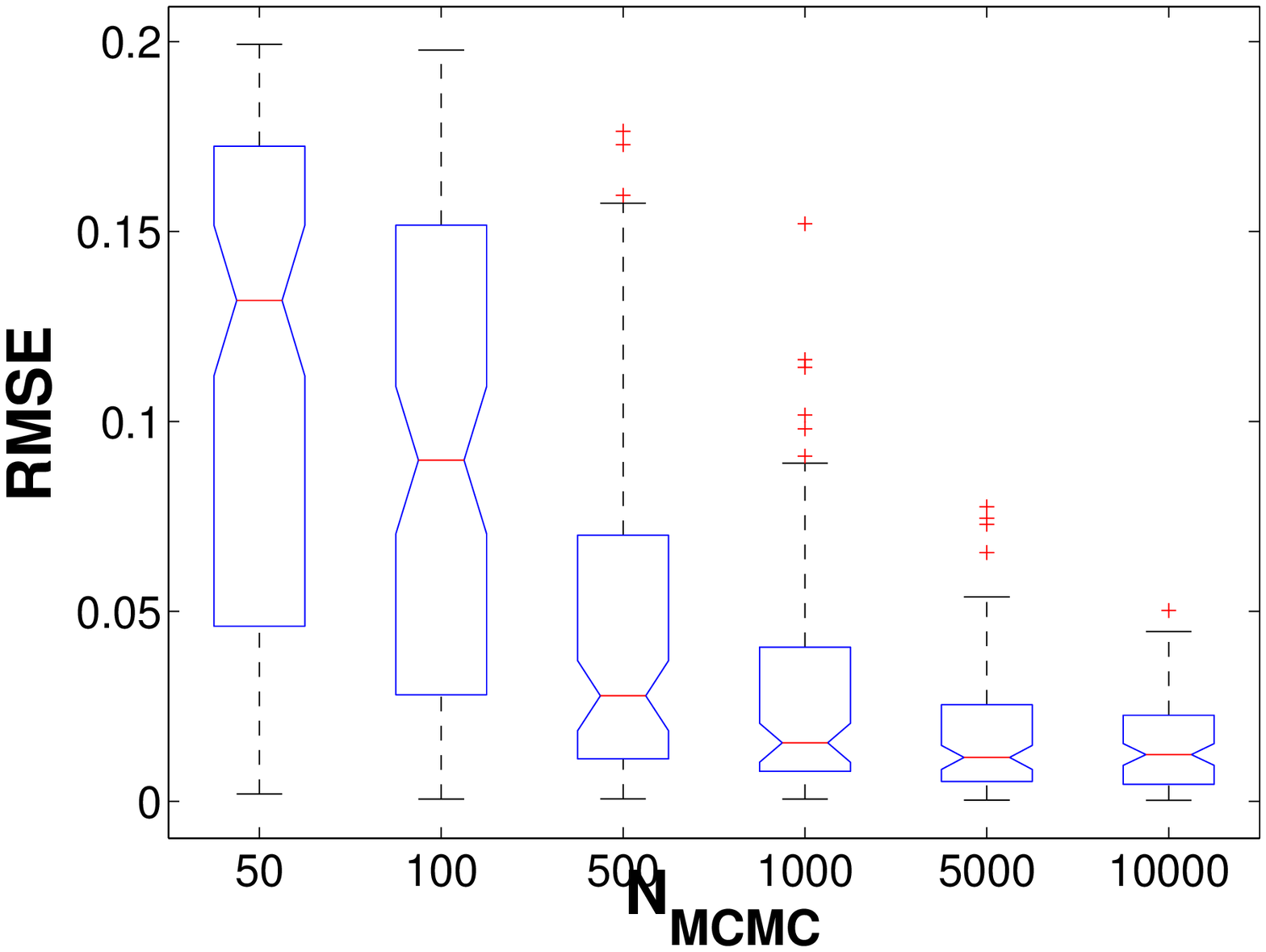} &
\includegraphics[scale=0.13]{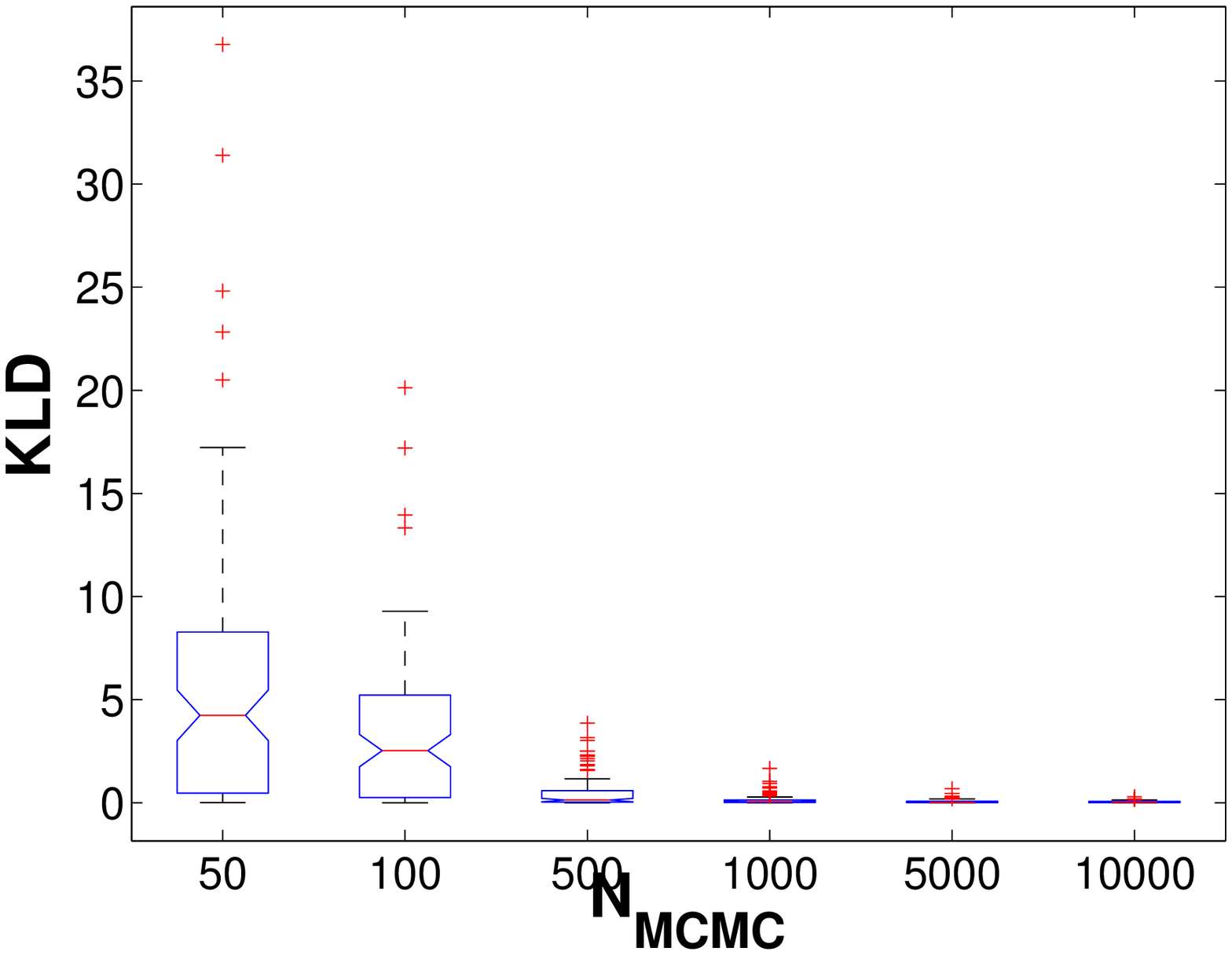} &
\includegraphics[scale=0.13]{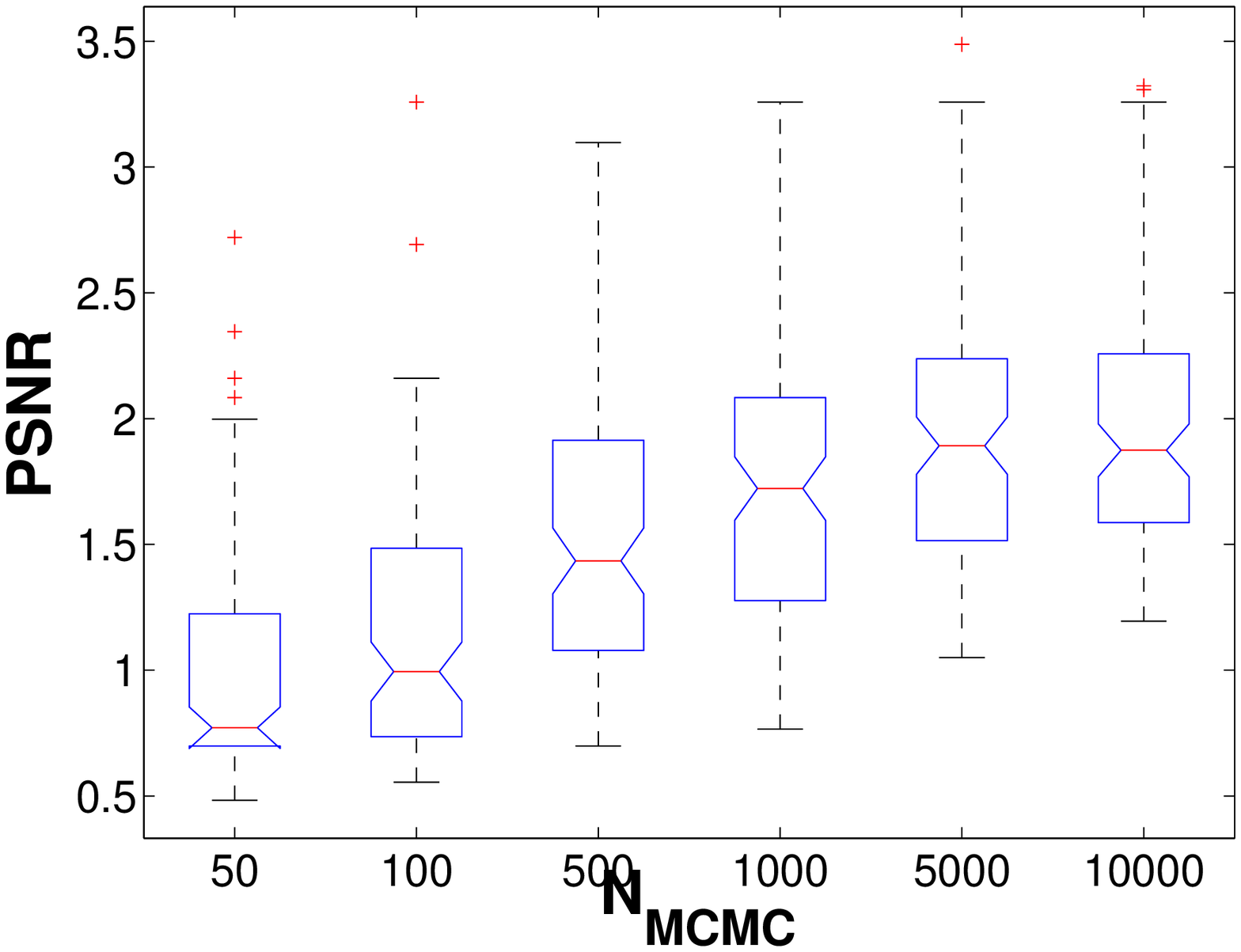} &
\includegraphics[scale=0.13]{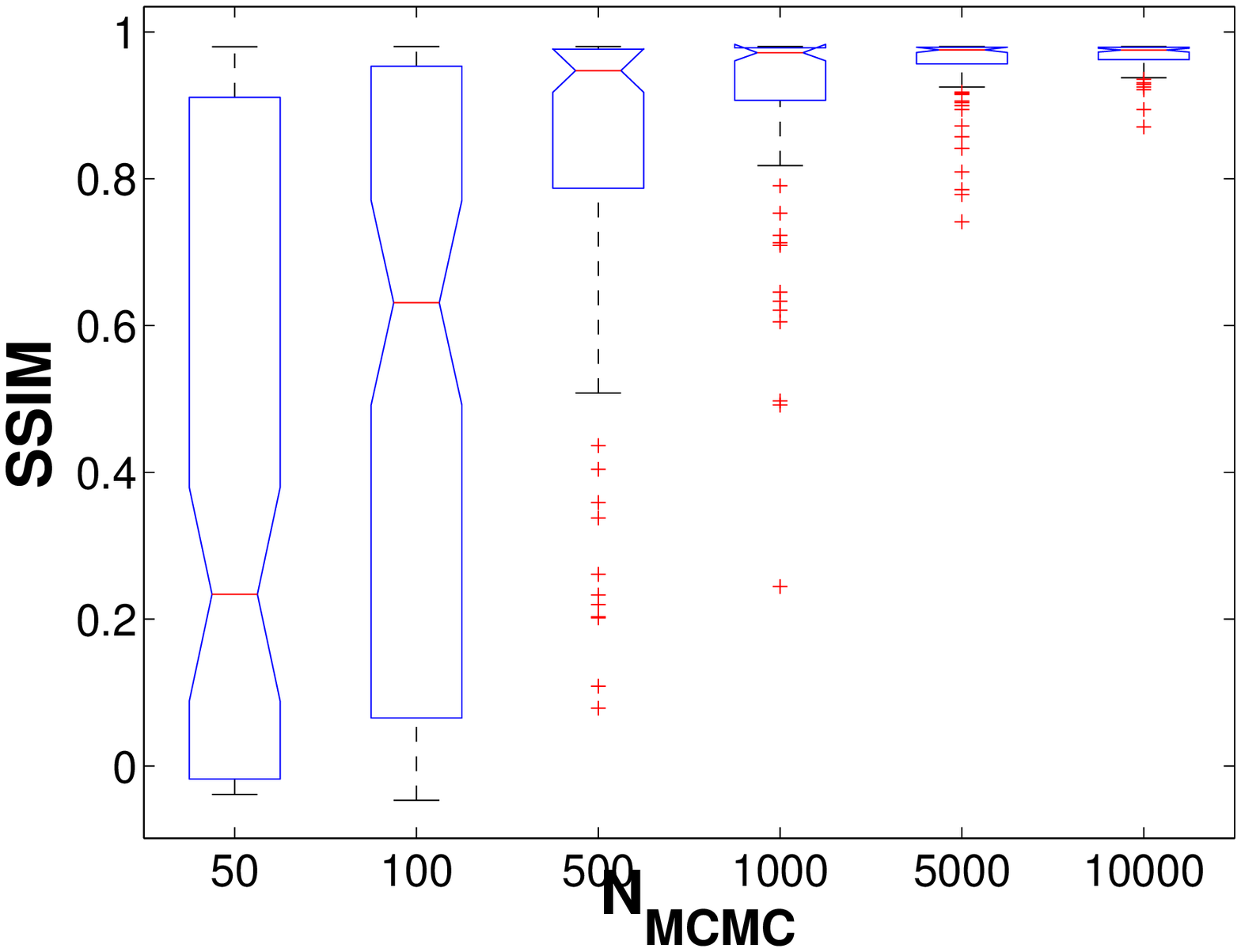} \cr
Fglass	&
\includegraphics[scale=0.13]{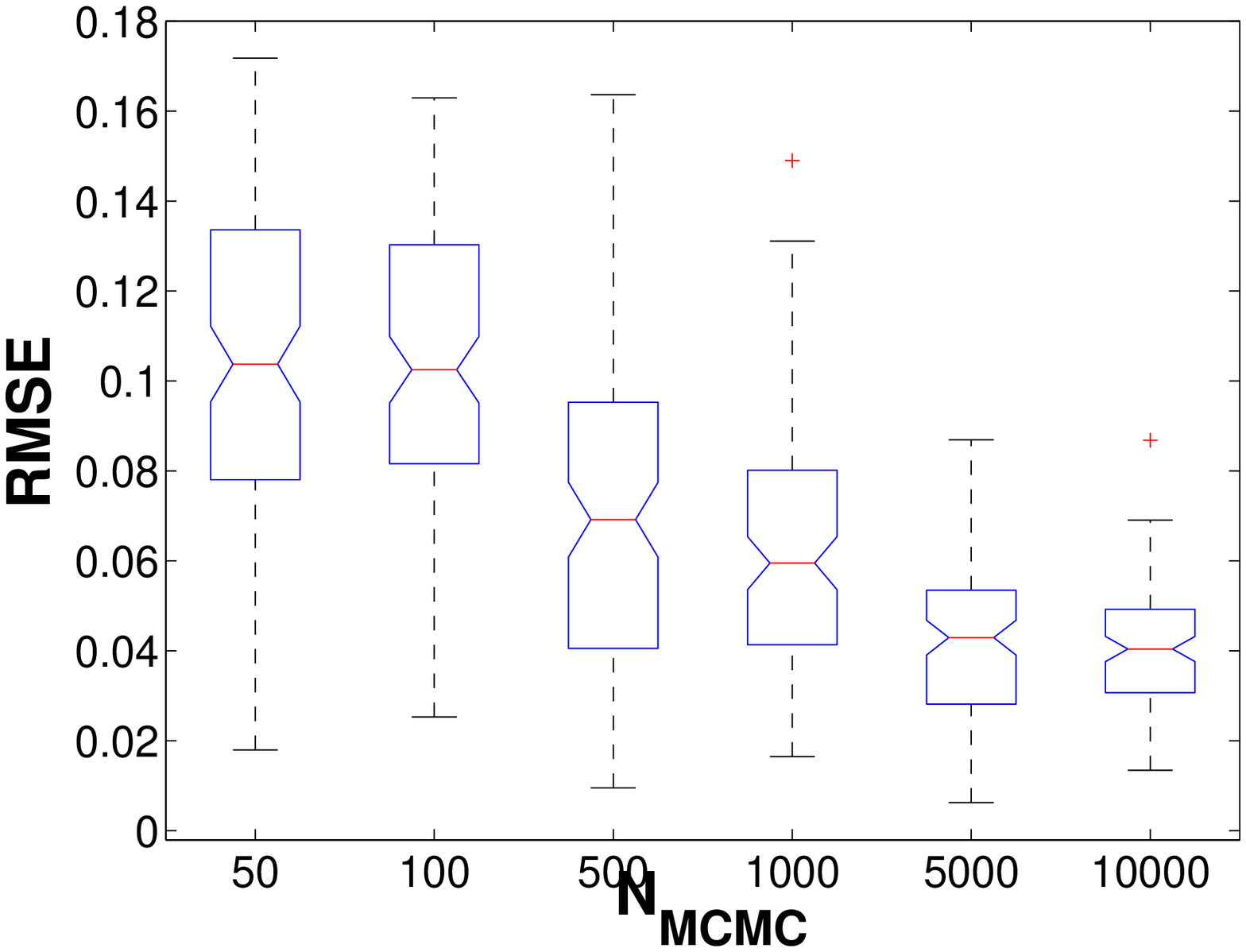} &
\includegraphics[scale=0.13]{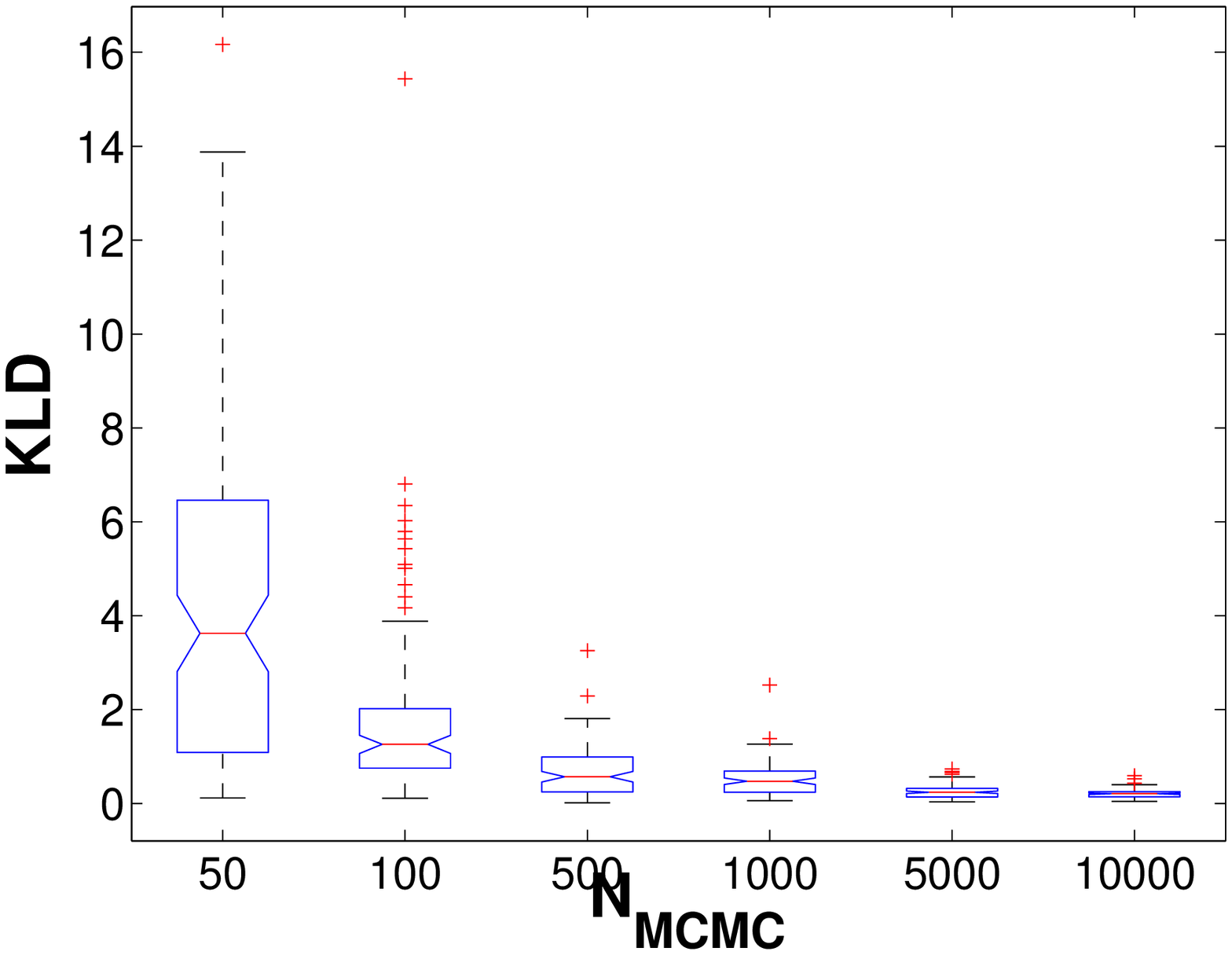} &
\includegraphics[scale=0.13]{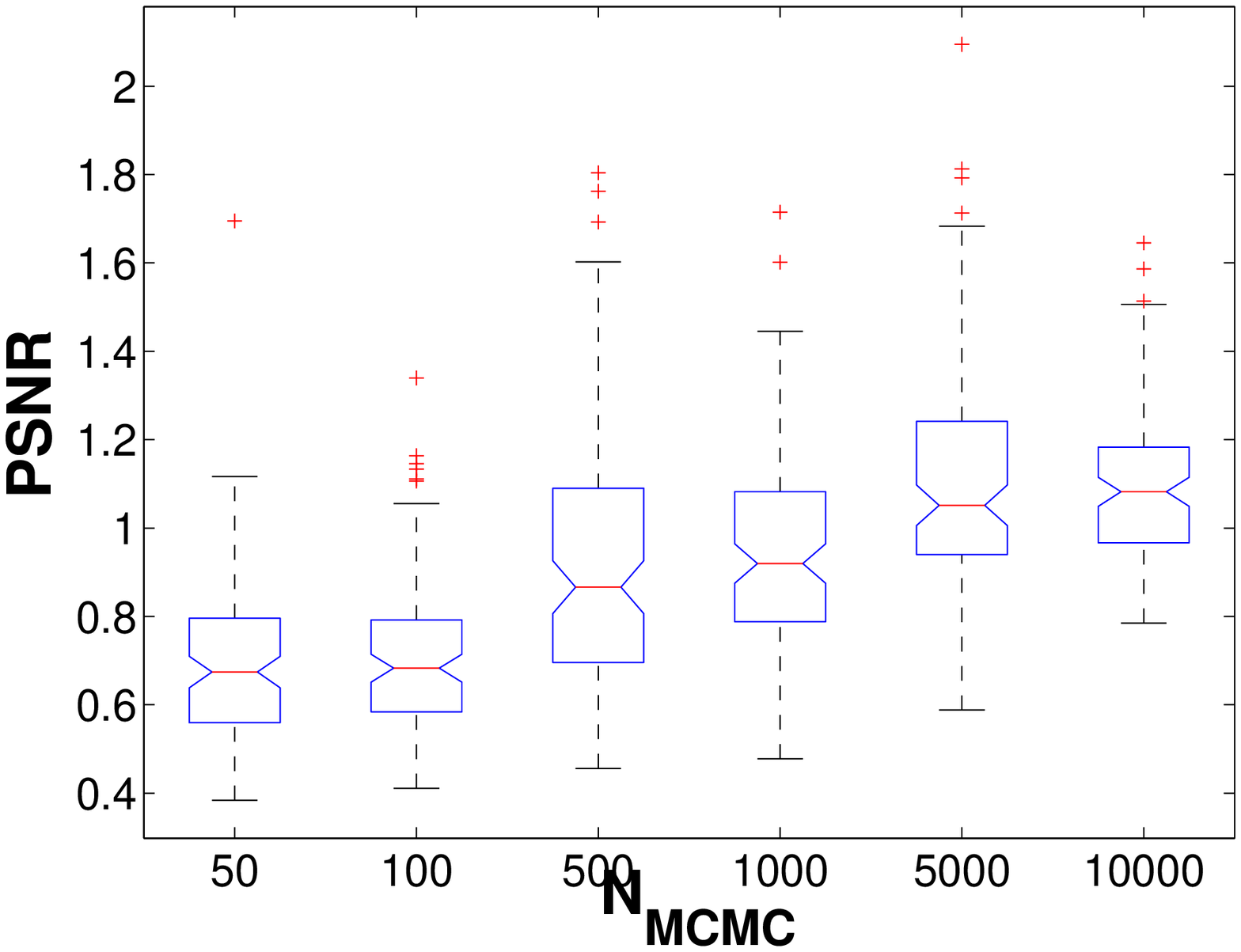} &
\includegraphics[scale=0.13]{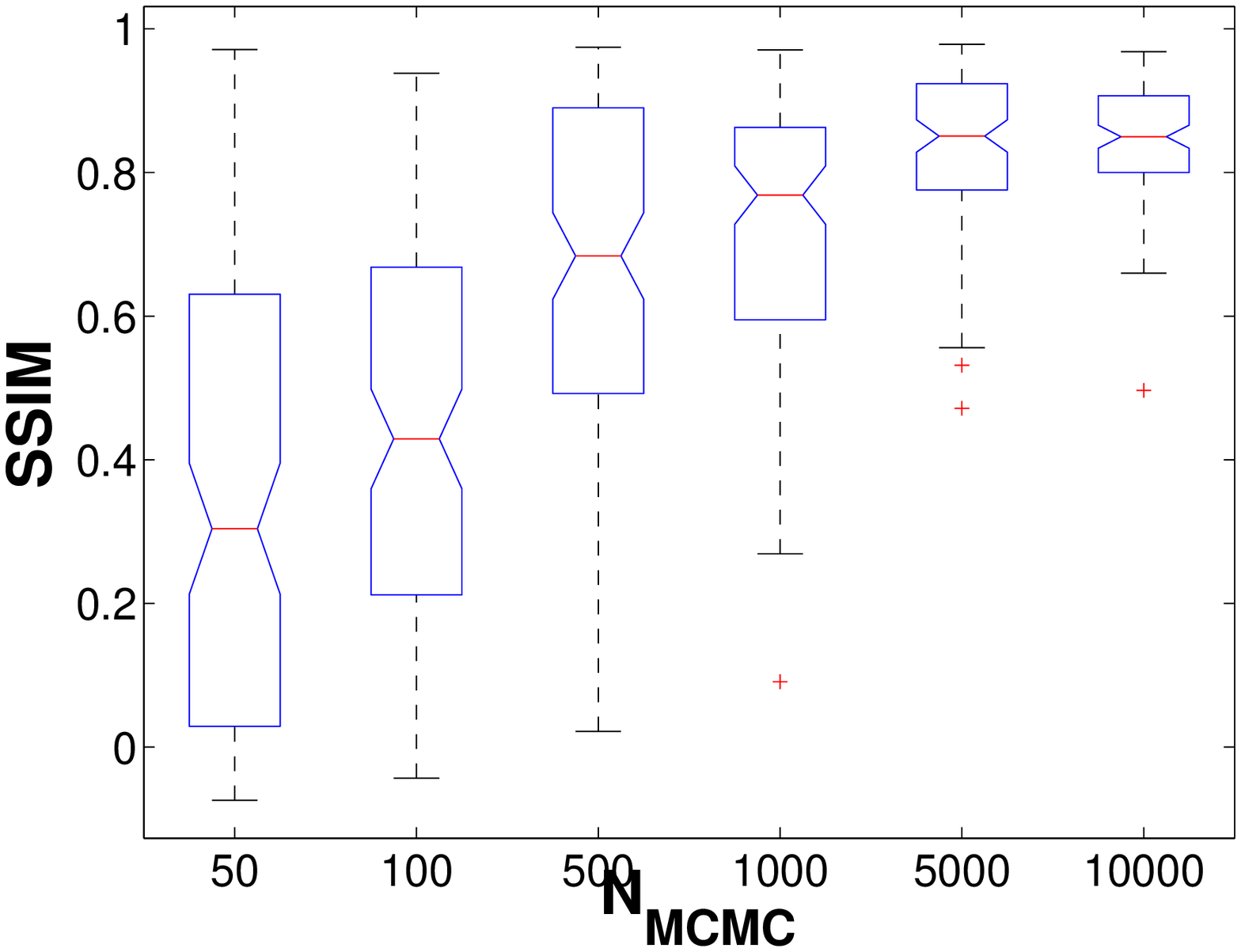} \cr
Meat		&
\includegraphics[scale=0.13]{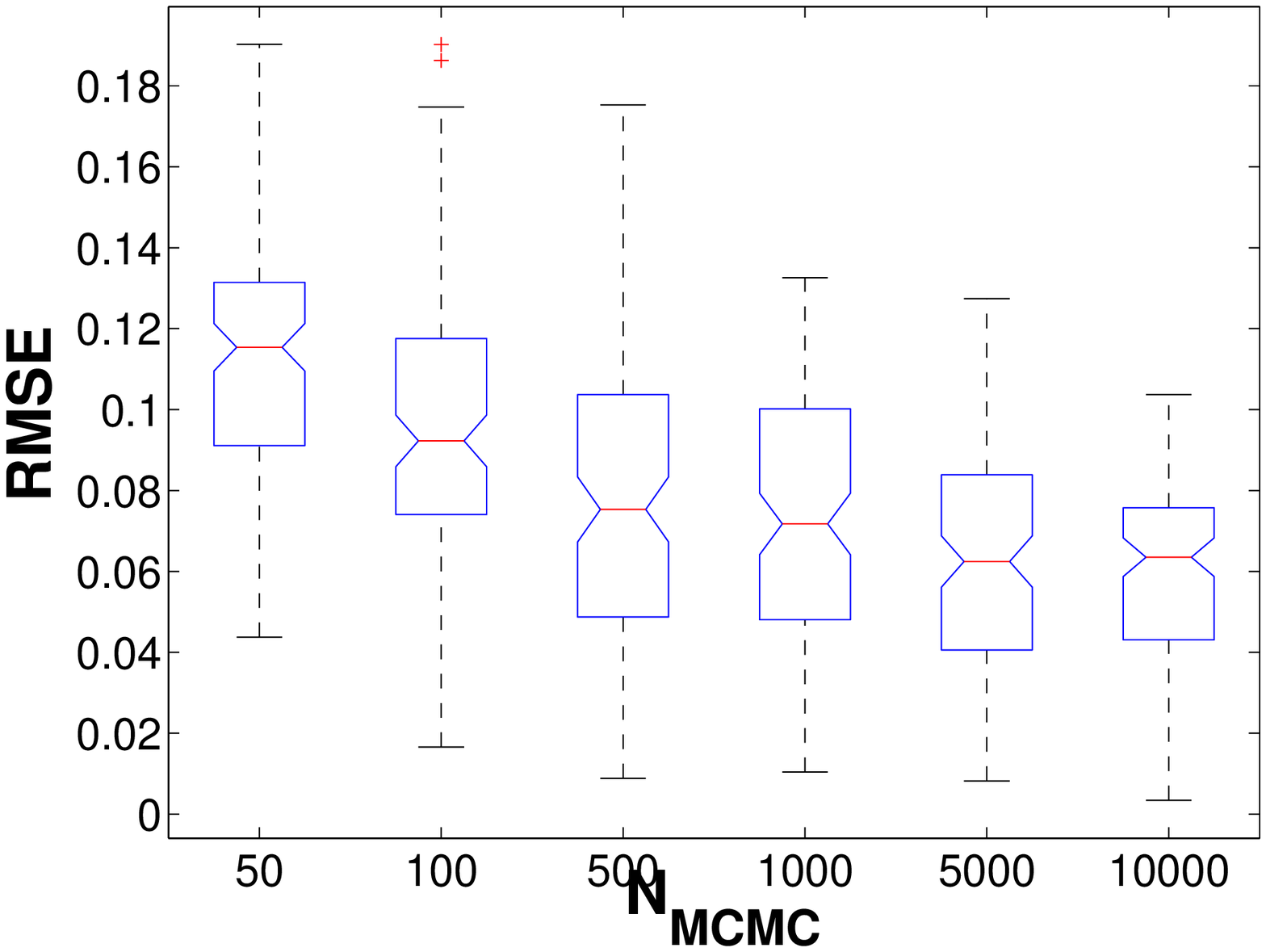} &
\includegraphics[scale=0.13]{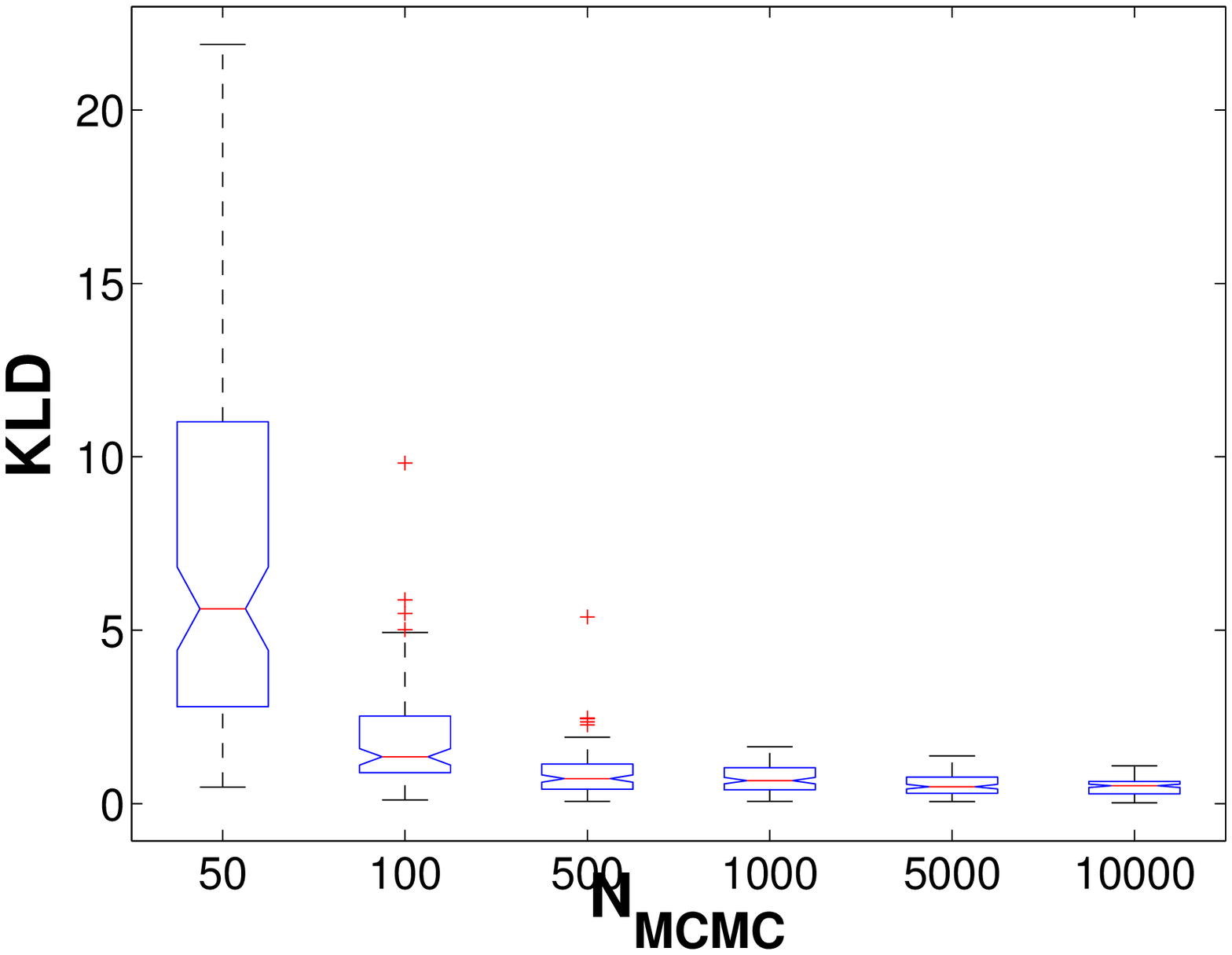} &
\includegraphics[scale=0.13]{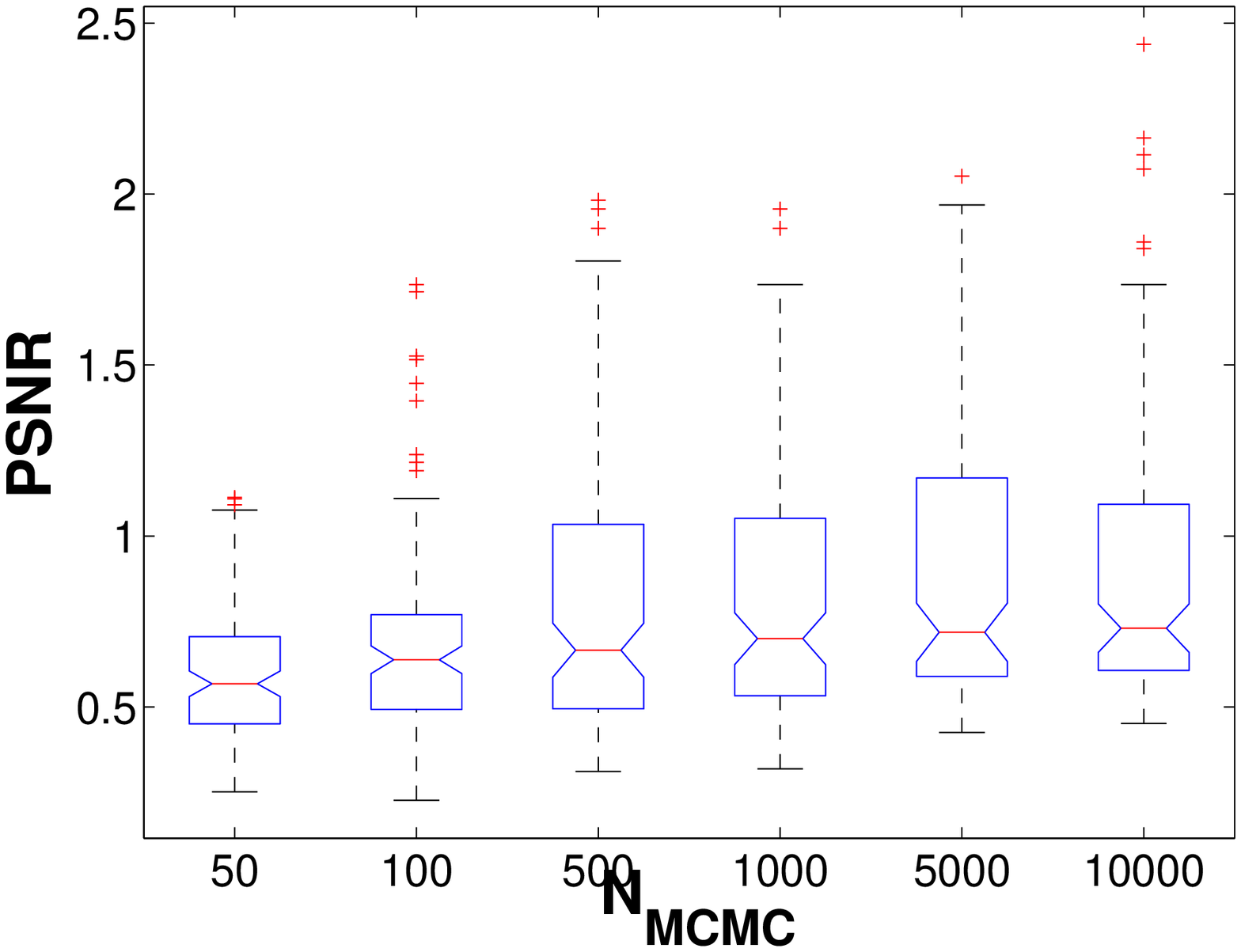} &
\includegraphics[scale=0.13]{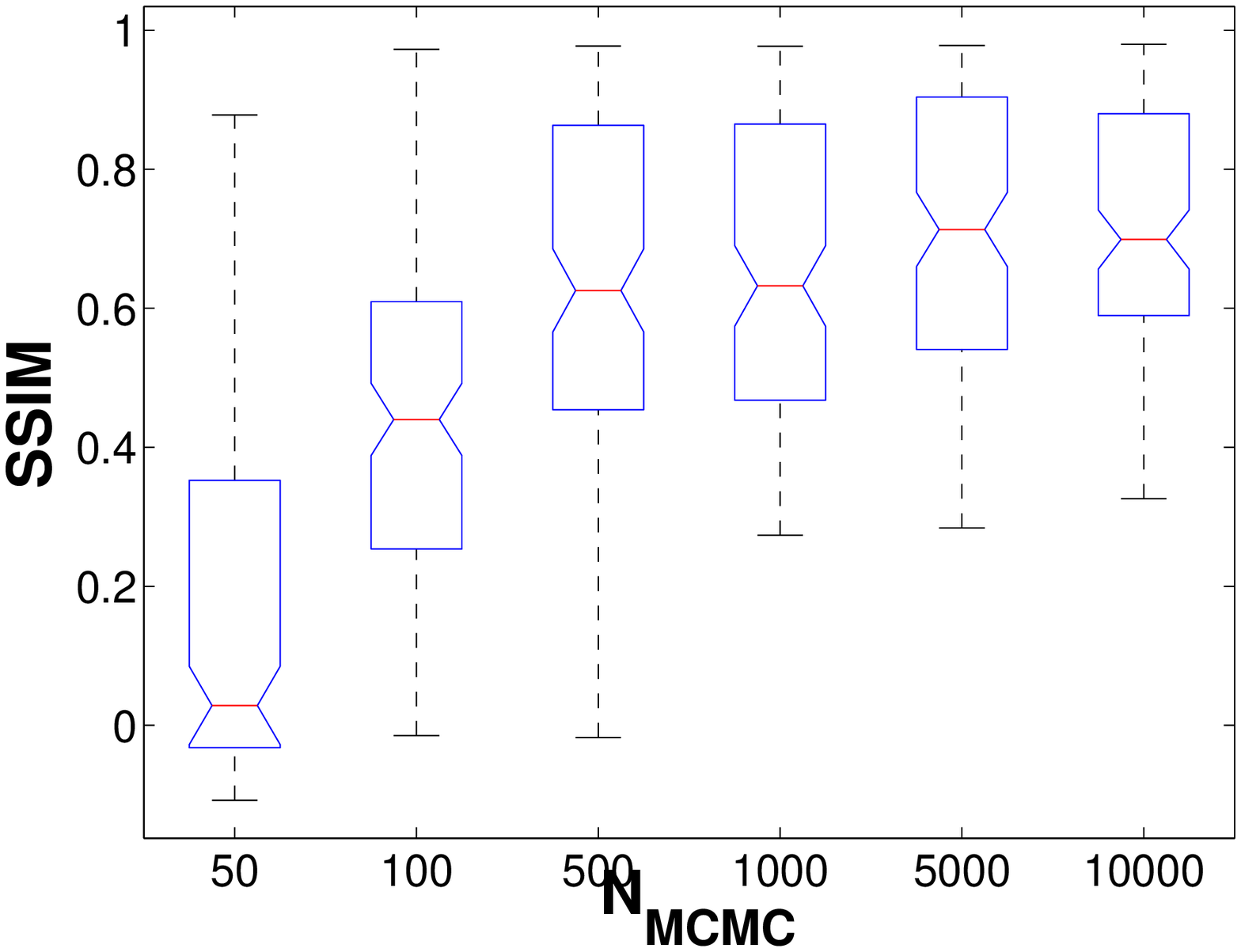} \cr
Oliveoil		&
\includegraphics[scale=0.13]{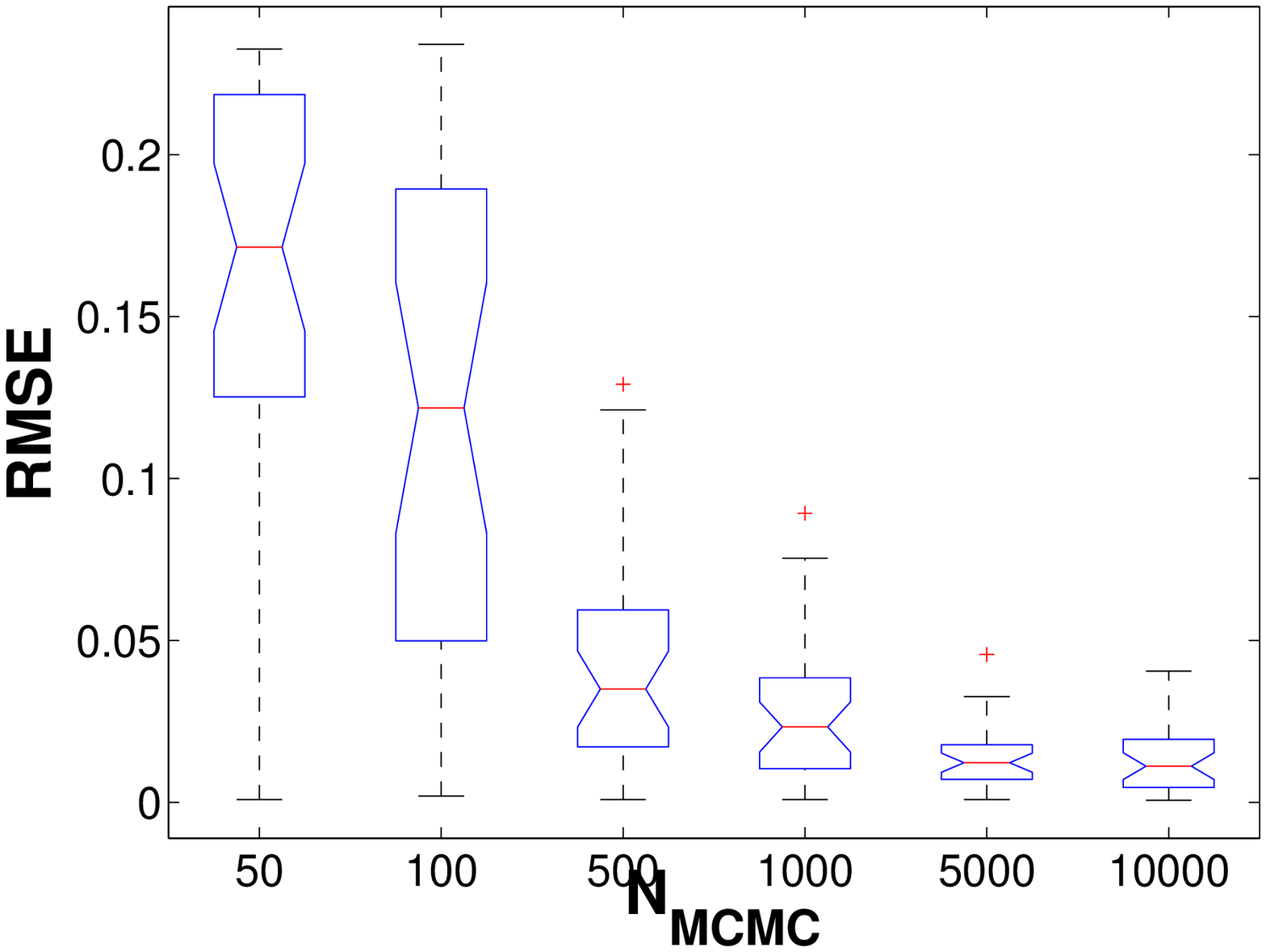} &
\includegraphics[scale=0.13]{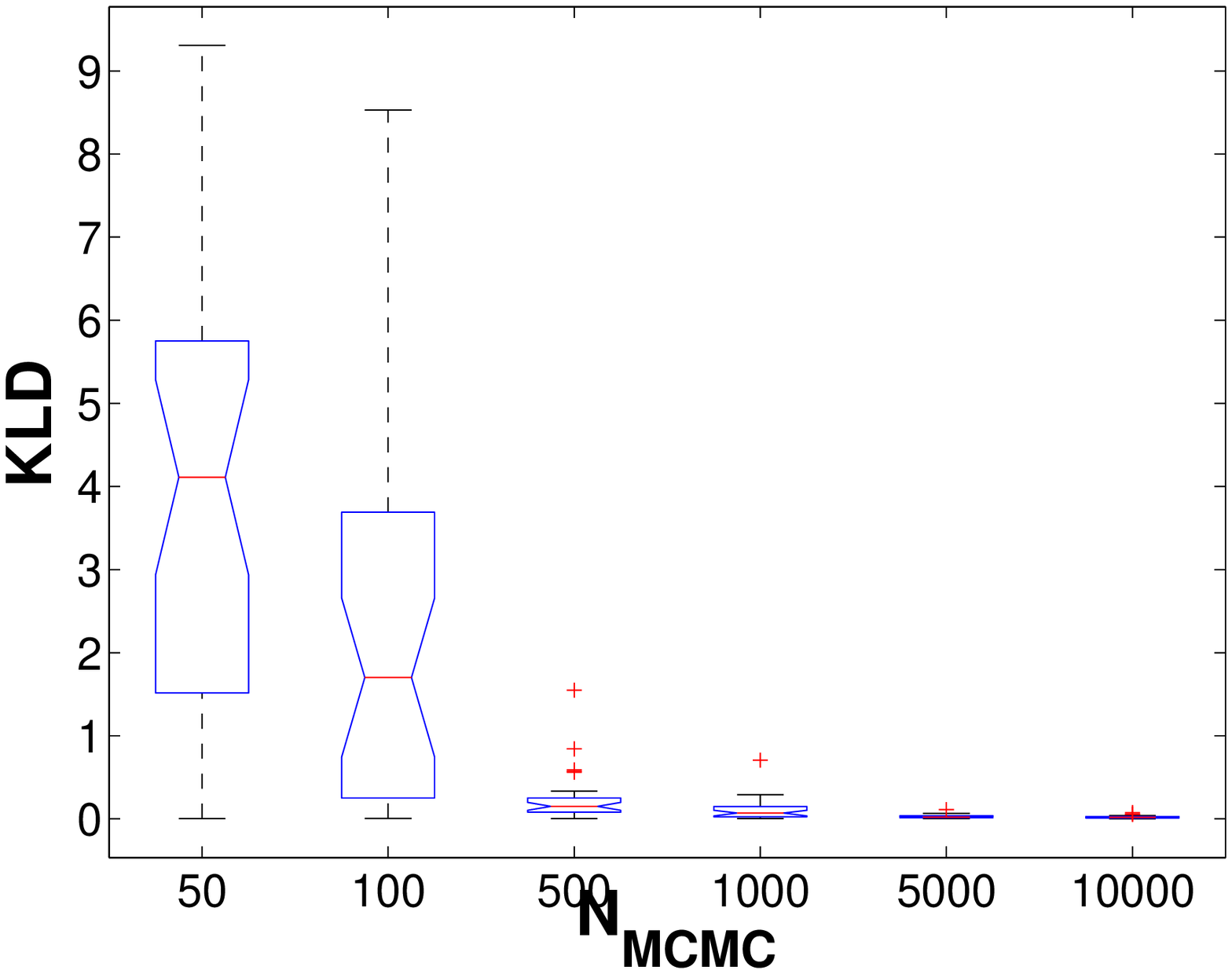} &
\includegraphics[scale=0.13]{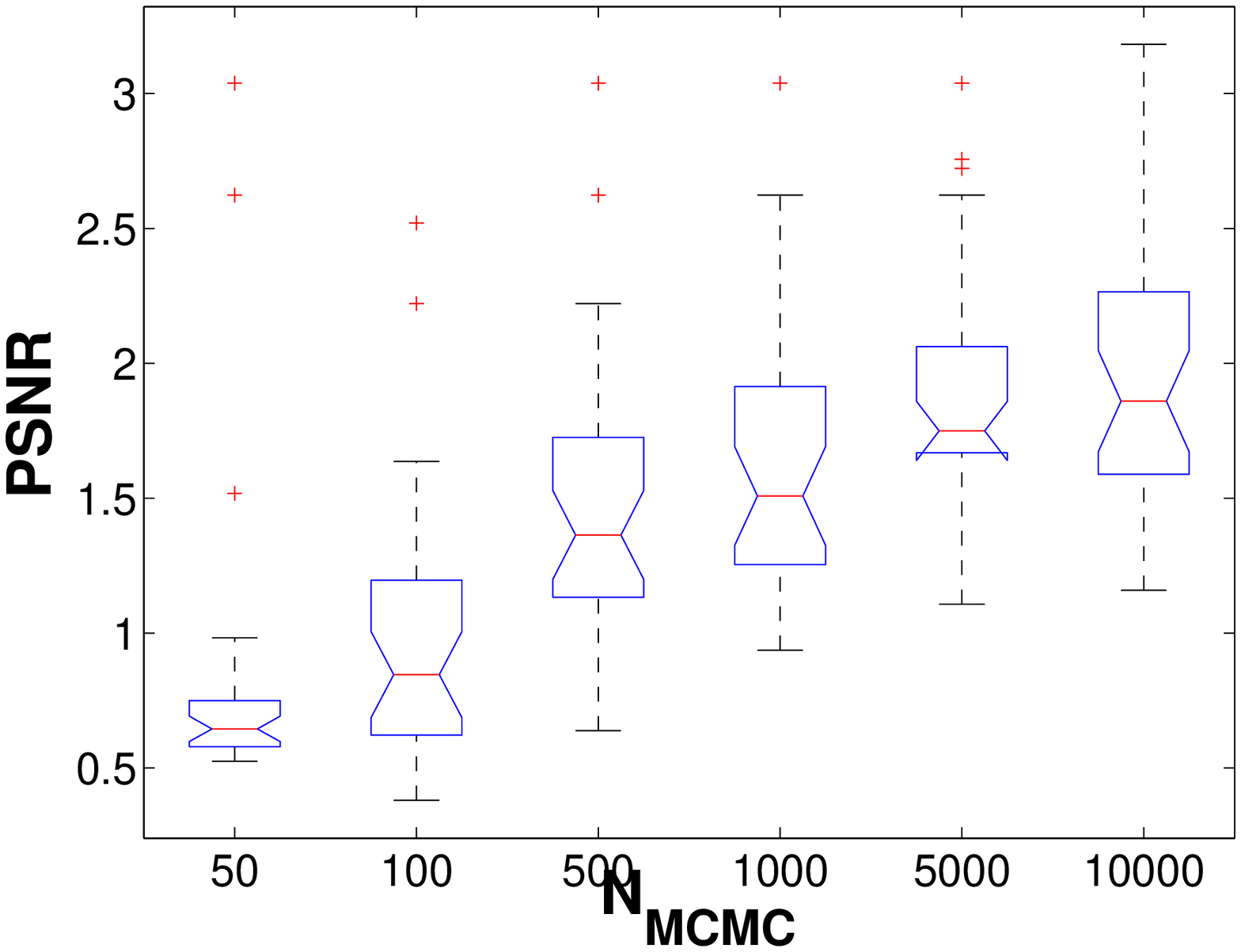} &
\includegraphics[scale=0.13]{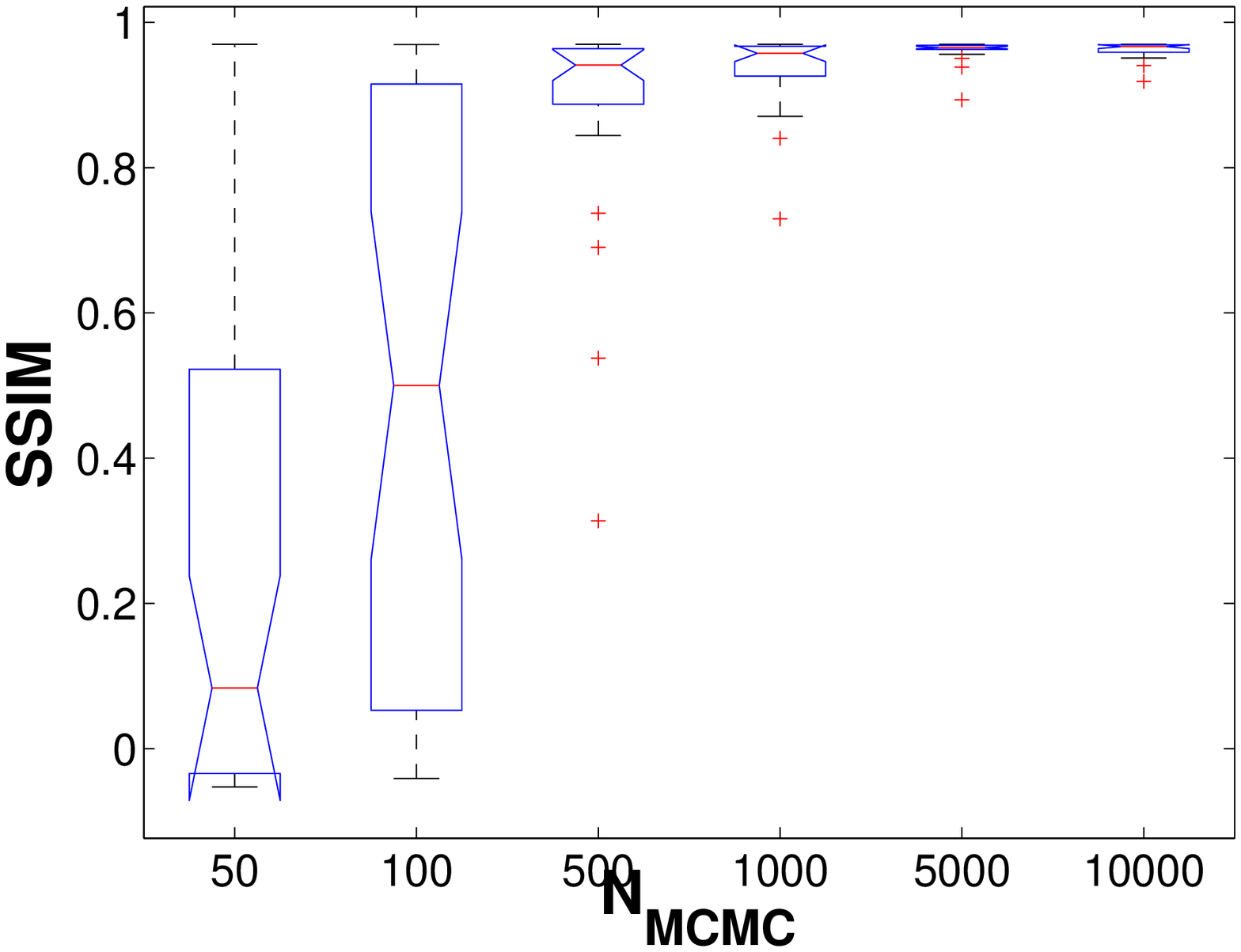} \cr
Wine		&
\includegraphics[scale=0.13]{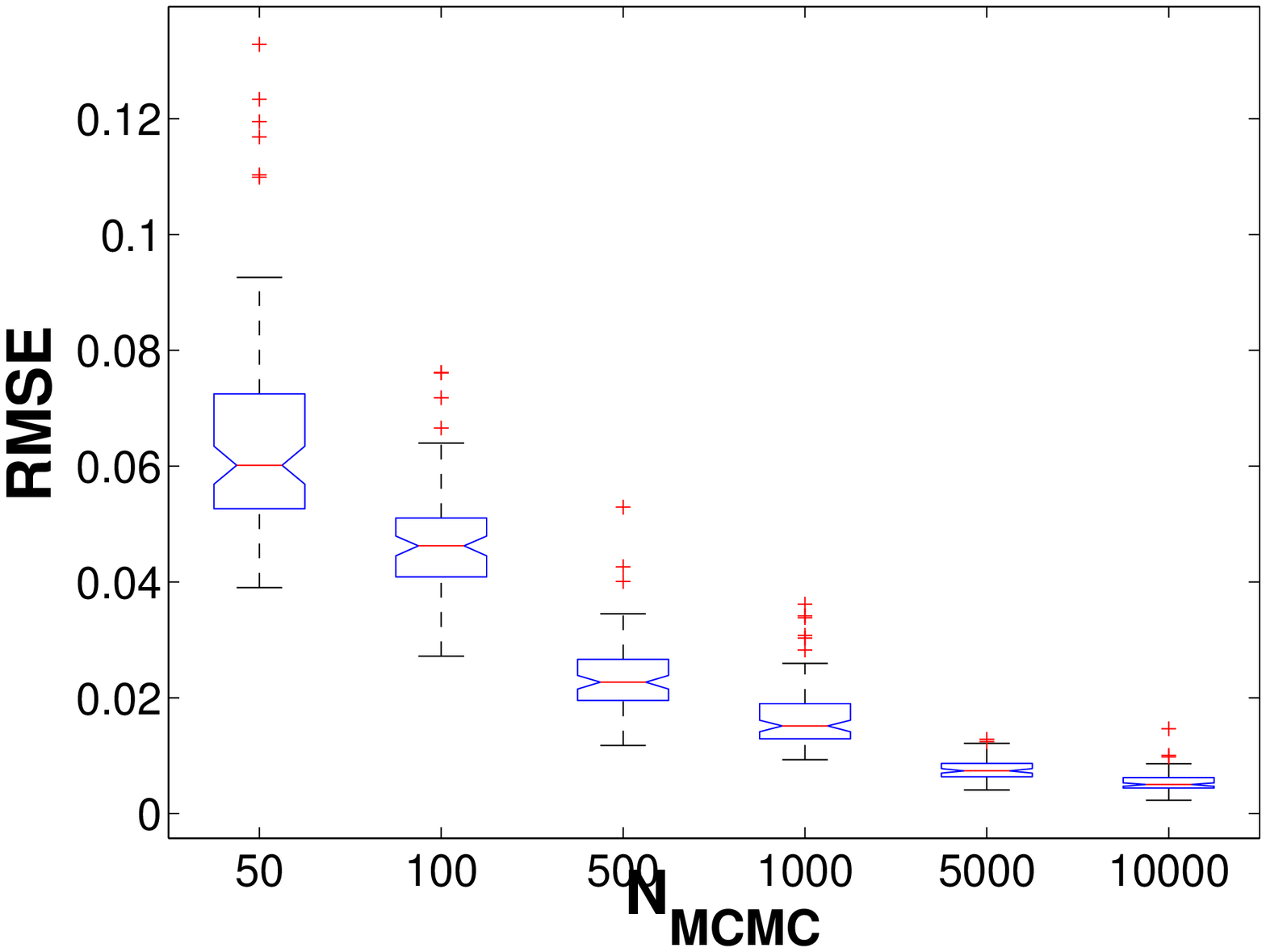} &
\includegraphics[scale=0.13]{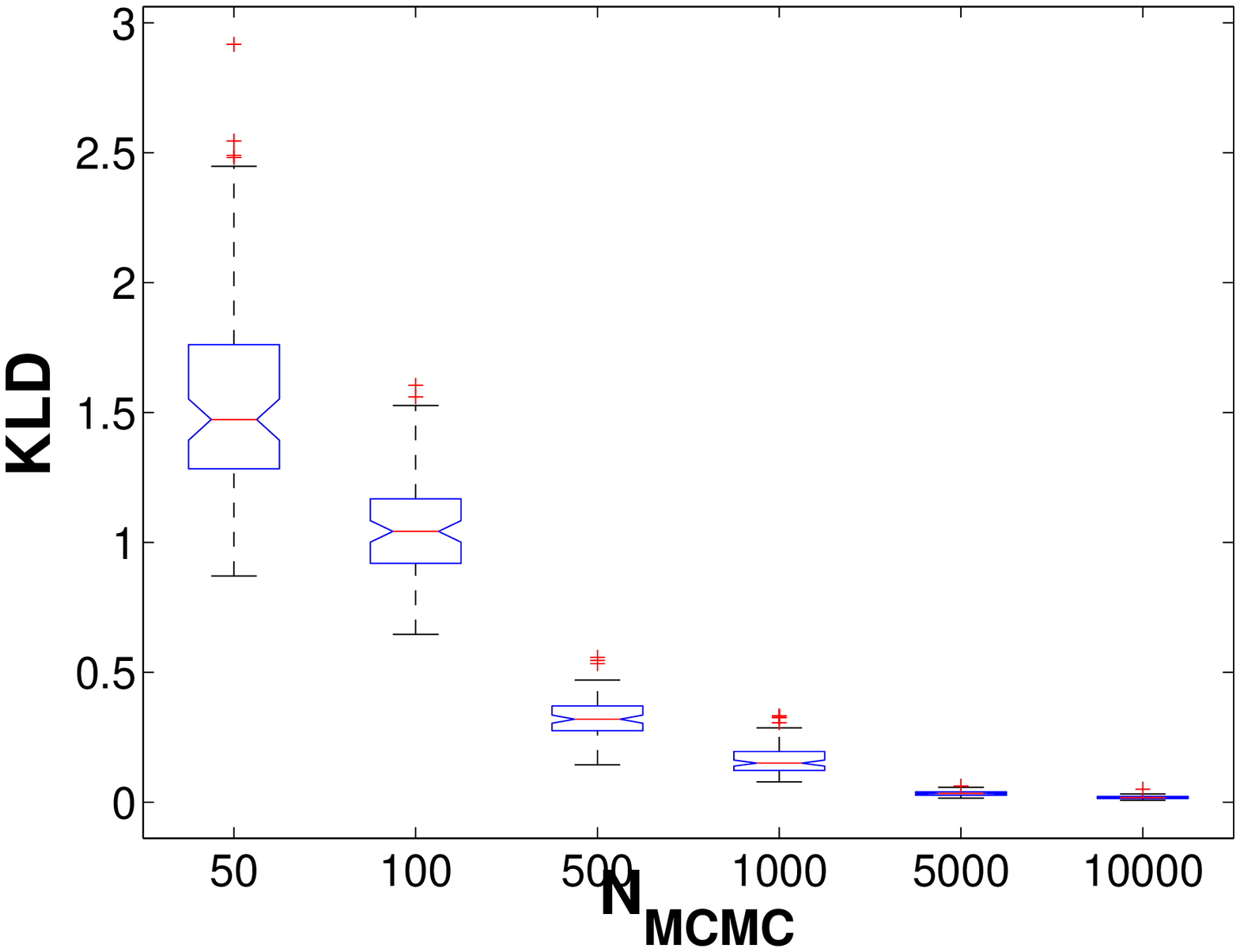} &
\includegraphics[scale=0.13]{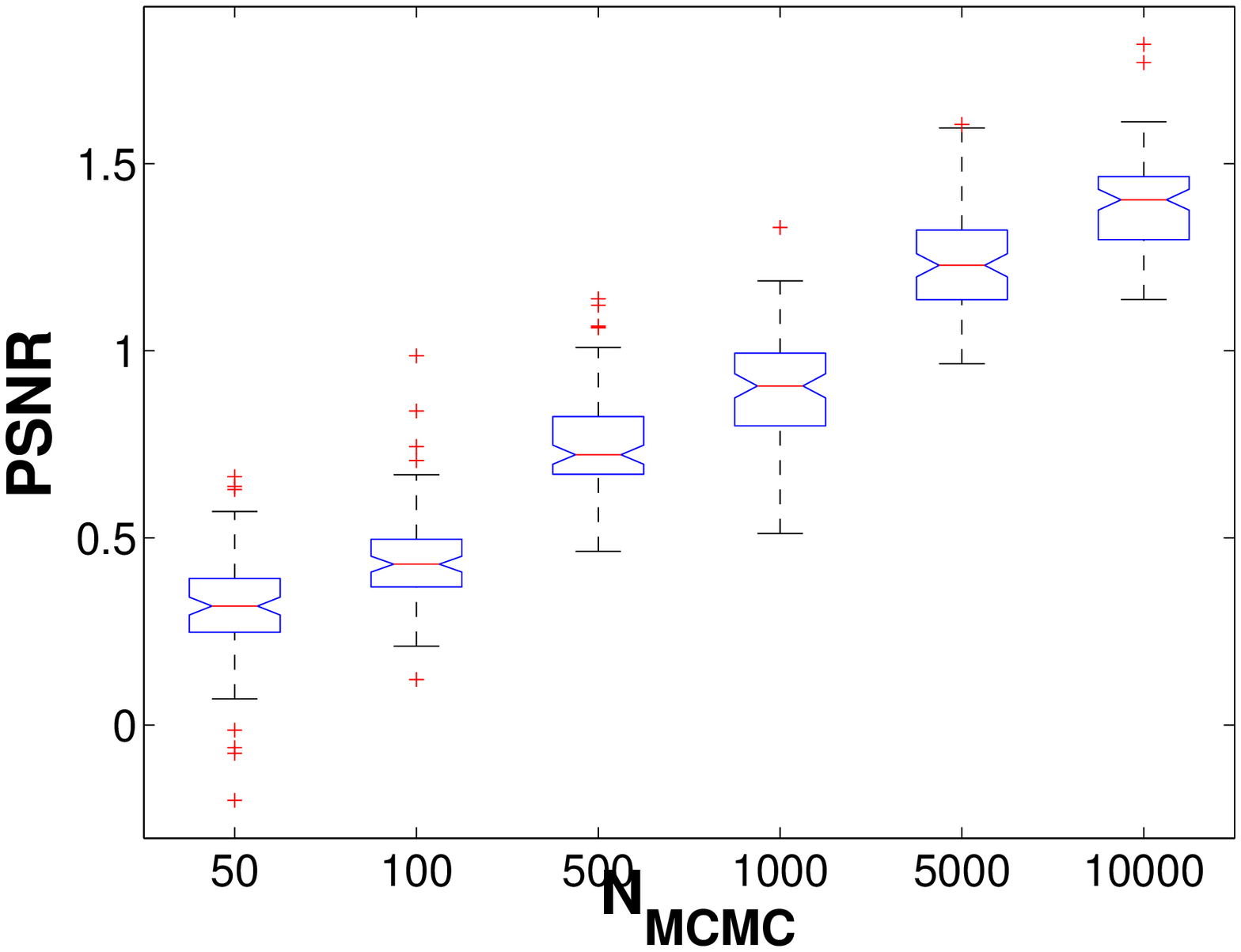} &
\includegraphics[scale=0.13]{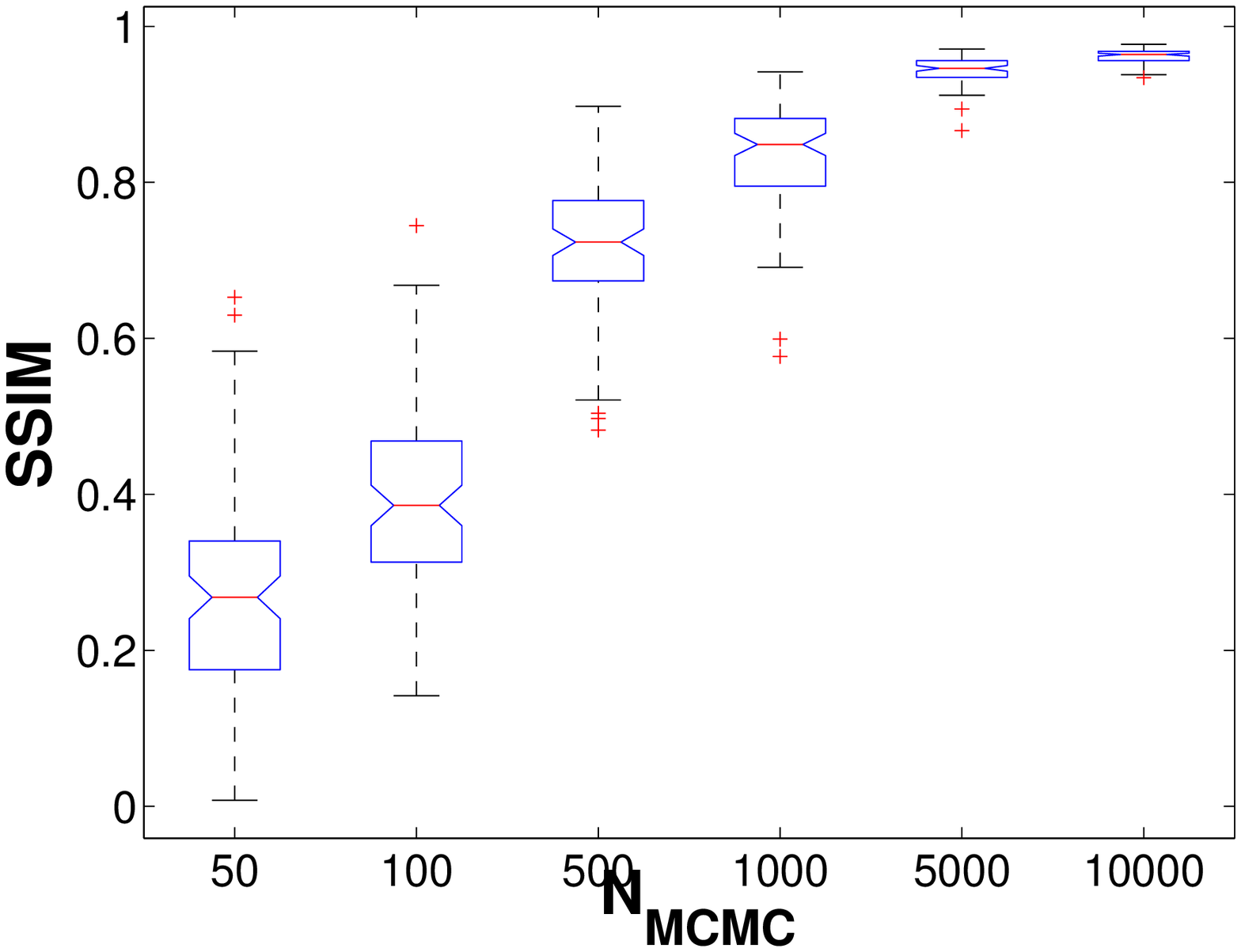} \cr
Iris 		&
\includegraphics[scale=0.13]{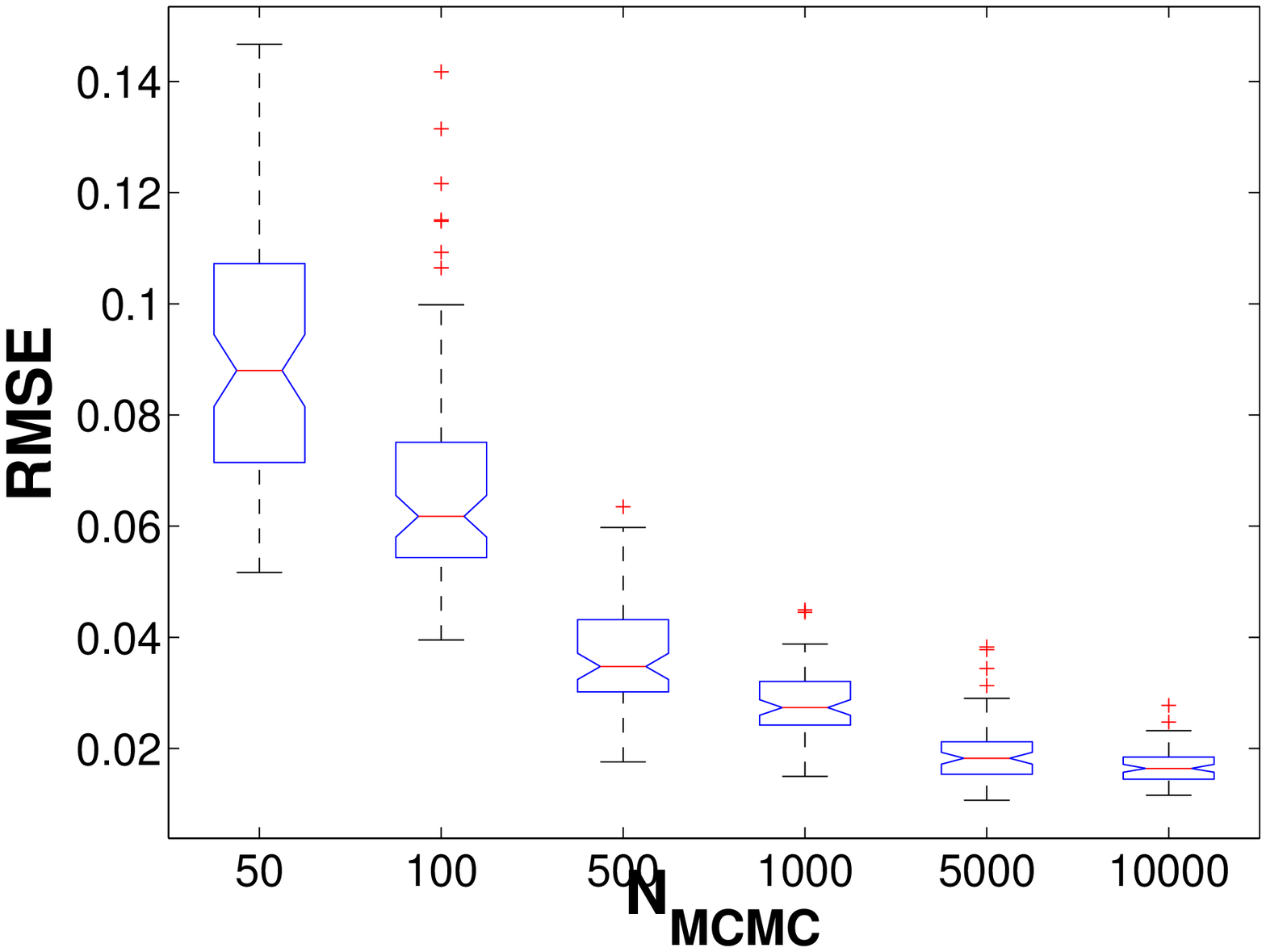} &
\includegraphics[scale=0.13]{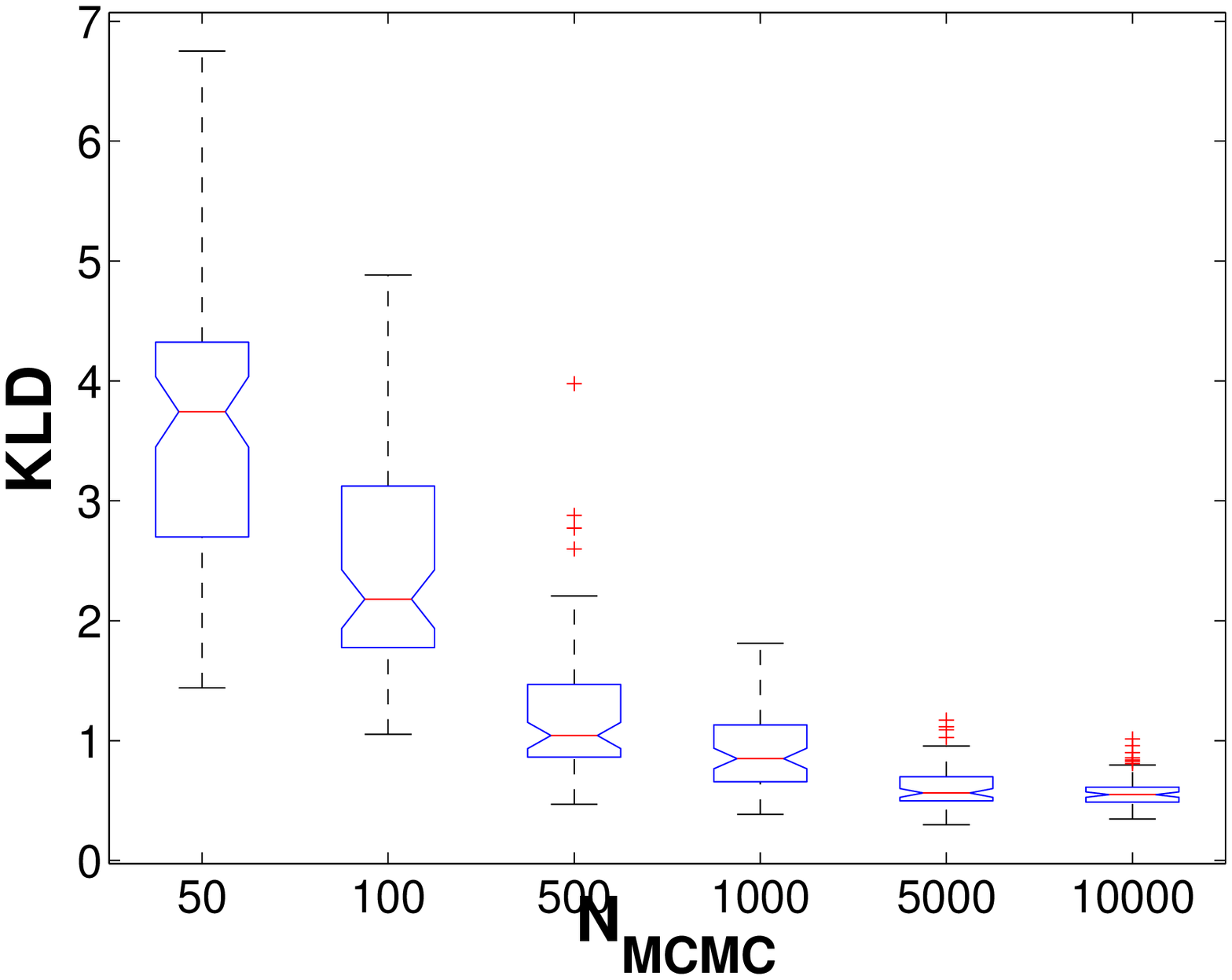} &
\includegraphics[scale=0.13]{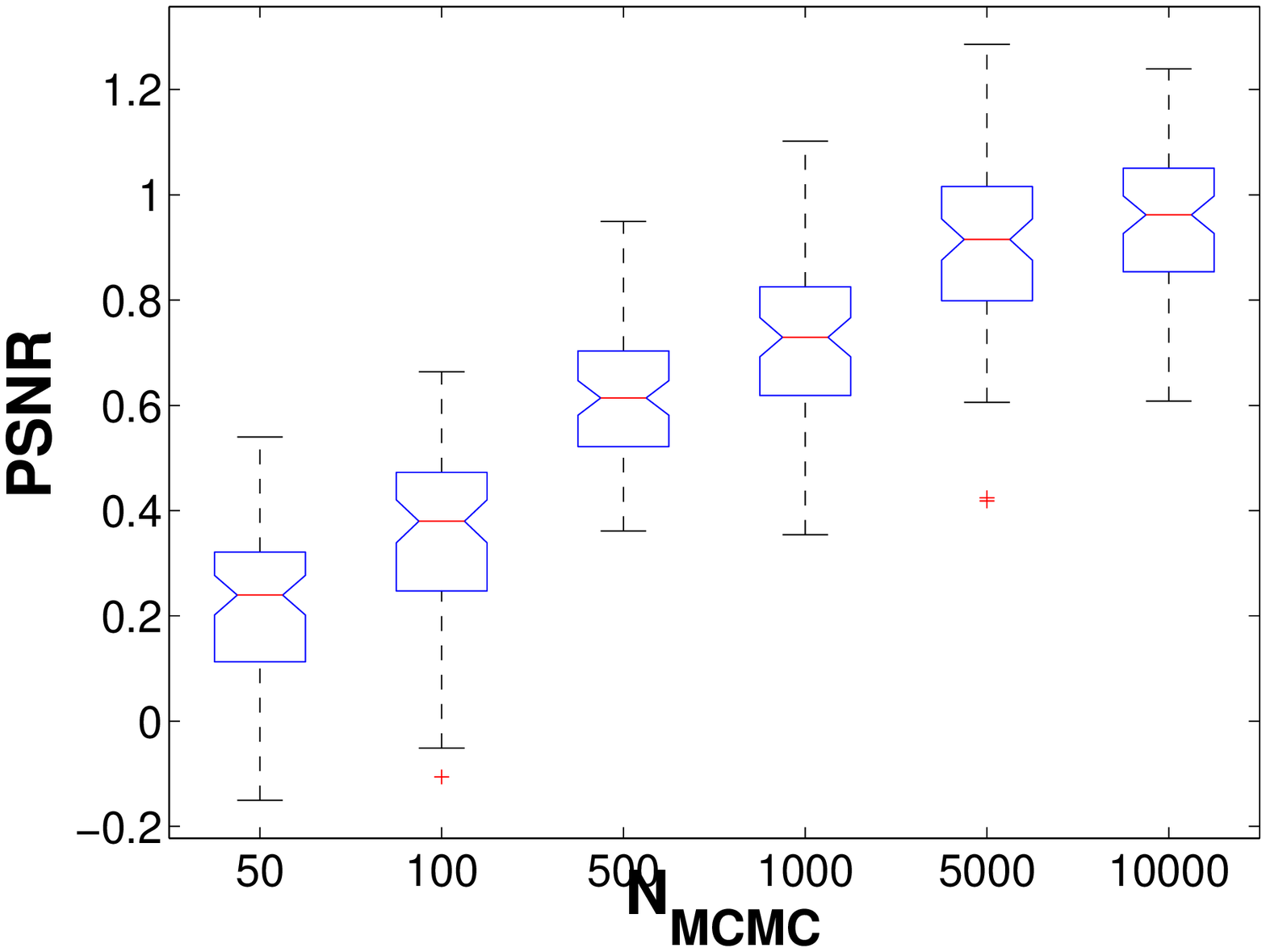} &
\includegraphics[scale=0.13]{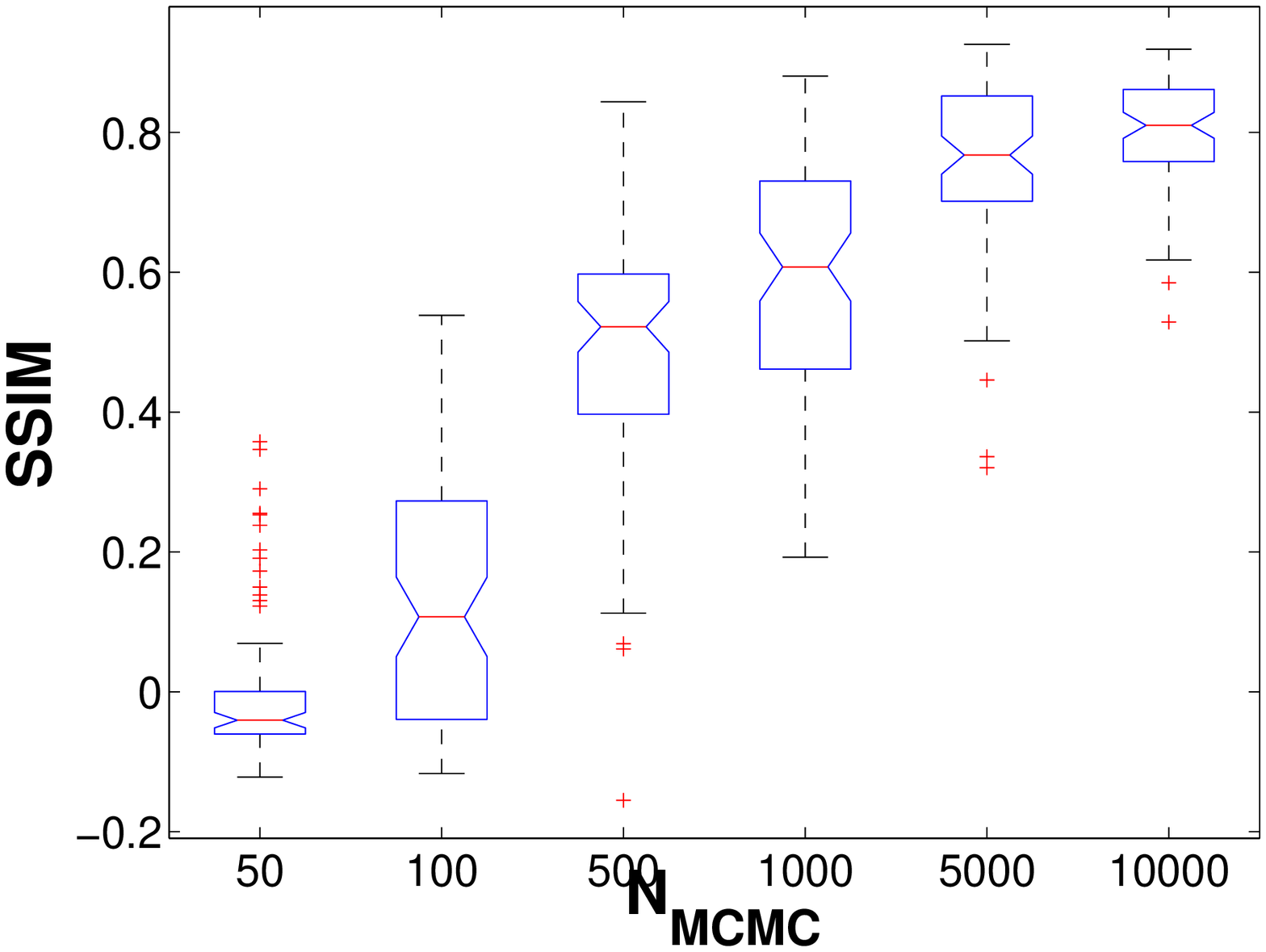}
\end{tabular}
\caption{Implicit similarity check between the reconstructed densities by MCMC and KOREA via four well-known metrics.}
\label{fig: Convergence checking for MCMC and KOREA}
\end{figure}

Asymptotically, MCMC with a large number of iterations will converge and therefore can be used in principle to estimate 
the underlying posterior density. Thus, we can check whether the reconstructed density using KOREA is close to that estimated
by MCMC with a very large number of iterations in order to validate the our proposed algorithm. Two subgraphs of 
figure \ref{fig: Comparison between MCMC and KOREA for wine dataset} visualize the similarity between reconstructed posterior 
densities of a testing data of wine dataset by KOREA (red circle line) and MCMC (blue cross line) with small (top) and large 
(bottom) number of samples. (For MCMC, we set the sample size by 100 for small size and 10000 for large size respectively.) 
As we can see in the figures, our propsed algorithm KOREA is closely approximated to the MCMC algorithm with a large number of
iterations ize which is 
commonly regarded as underlying reference or pseudo-ground truth density.
In order to measure the similarity between the reconstructed densities by MCMC and KOREA, we use four different metrics as shown 
in figure \ref{fig: Convergence checking for MCMC and KOREA}: Root Mean Square Error (RMSE), Peak Signal-to-Noise Ratio (PSNR), 
Kullback Leibler Distance (KLD) and Structural SIMilarity (SSIM) \cite{Wang04imagequality}. As in the case of 
figure \ref{fig: Comparison between MCMC and KOREA for wine dataset}, MCMC with a large sample size produces densities very close 
to those produced by our proposed KOREA algorithm. As the number of MCMC samples increases, RMSE and  KLD decrease while PSNR and 
SSIM increases for all datasets.

Table \ref{table: Comparison of F-measures of Benchmark datasets} demonstrates the performance of the each algorithms based on 
F-measure for four cases: kNN, PKNN, KOREA (average) and KOREA (optimal). Since MCMC produces results which are very close to 
that of KOREA as shown in figures \ref{fig: Comparison between MCMC and KOREA for wine dataset} and 
\ref{fig: Convergence checking for MCMC and KOREA}, we did not present these results. KOREA (average) and KOREA (optimal) 
represent the mean (marginalized) estimate and MAP estimate of KOREA, respectively. As we can see in the table, KOREA works 
superior to other conventional approaches for all datasets. The results with the best performance are highlighted in bold in 
this table. 
\begin{table}[h!]
\caption{Comparison of F-measures with varying neighbouring structures. The results with the best performance are written in bold.}
\label{table: Comparison of F-measures of Benchmark datasets}
\centering
\begin{tabular}{||c|c|c|c|c||}
\hline\hline
Methods	 	& Data	 & Asymmetric 	& Symmetric 		 & Boltzman$^{(2)}$   \cr
		 	& 		 & model 		& Boltzman 		 &   \cr
\hline
\multirow{6}{*}{KNN}			&Crabs	&0.72$\pm$0.08		& 0.74$\pm$0.08		&0.74$\pm$0.08			\cr
							&Fglass	&0.64$\pm$0.06		&0.67$\pm$0.05			&0.67$\pm$0.05			\cr
					 		&Meat	&0.68$\pm$0.07		&0.70$\pm$0.07			&0.70$\pm$0.07			\cr
							&Oliveoil	&0.74$\pm$0.12		&0.71$\pm$0.10			&0.71$\pm$0.10			\cr
							&Wine	&0.97$\pm$0.01		&0.97$\pm$0.01 		&0.97$\pm$0.01			\cr
							&Iris		&0.57$\pm$0.10		& 0.55$\pm$0.10		&0.55$\pm$0.10			\cr
\hline
\multirow{6}{*}{PKNN}			&Crabs	& 0.75$\pm$0.09	&0.75$\pm$0.09			&0.75$\pm$0.08			\cr
							&Fglass	&0.73$\pm$0.06		&0.74$\pm$0.06			&0.69$\pm$0.06			\cr
							&Meat	&0.70$\pm$0.07		&0.71$\pm$0.07			&0.70$\pm$0.06			\cr
							&Oliveoil	&0.72$\pm$0.11		&0.73$\pm$0.11			&0.70$\pm$0.11			\cr
							&Wine	&0.98$\pm$0.01		&0.98$\pm$0.01			&0.98$\pm$0.02			\cr
							&Iris		& 0.57$\pm$0.12	&0.57$\pm$0.12			&0.53$\pm$0.10			\cr
\hline
	 						&Crabs	& 0.86$\pm$0.11	&{\bf 0.89$\pm$0.11}			&0.87$\pm$0.09			\cr
							&Fglass	&0.76$\pm$0.09		&0.77$\pm$0.07			&{\bf 0.81$\pm$0.08}	\cr
KOREA						&Meat	&0.68$\pm$0.12		&{\bf 0.75$\pm$0.06}	&0.71$\pm$0.06			\cr
(average)						&Oliveoil	&{\bf 0.82$\pm$0.17}		&0.76$\pm$0.19			&0.73$\pm$0.20			\cr
							&Wine	&0.99$\pm$0.12		&{\bf 0.99$\pm$0.02}	&0.98$\pm$0.02			\cr
							&Iris		&{\bf 0.62$\pm$0.15}	&0.58$\pm$0.17			&0.56$\pm$0.03			\cr
\hline
	 						&Crabs	&0.86$\pm$0.13		&{\bf 0.89$\pm$0.11}			&0.87$\pm$0.09			\cr
							&Fglass	&0.79$\pm$0.04		&0.76$\pm$0.08			&0.79$\pm$0.07			\cr
KOREA						&Meat	&0.70$\pm$0.11		&0.73$\pm$0.07			&0.69$\pm$0.13			\cr
(optimal)						&Oliveoil	&0.80$\pm$0.17		&0.78$\pm$0.17			&0.76$\pm$0.19			\cr
							&Wine	&{\bf 0.99$\pm$0.02} 	&{\bf 0.99$\pm$0.02}			&0.98$\pm$0.02			\cr
							&Iris		&0.57$\pm$0.17		&0.56$\pm$0.19			&0.48$\pm$0.16			\cr
\hline\hline
\end{tabular}
\end{table}

In addition, we compared the simulation times for each of the algorithms. Table \ref{table: Time comparison} demonstrates the 
execution time for all algorithms. Our proposed algorithm (PKNN with KOREA) is slower than conventional kNN and PKNN with 
fixed $K$ but it is much faster than MCMC technique which is regarded as one of the best approaches to infer the model parameters 
and number of neighbours in Bayesian framework. From the point of the accuracy of table 
\ref{table: Comparison of F-measures of Benchmark datasets} and the execution time of table \ref{table: Time comparison}, we 
eventually find that PKNN can be efficiently improved by using our proposed KOREA algorithm and this is a very practically useful 
technique compared to the conventional approaches including KNN, PKNN and MCMC. 
\begin{table}[h!]
\caption{Time comparison: the average of the execution times}
\label{table: Time comparison}
\centering
\begin{tabular}{c|c|c|c|g}
\hline\hline
Data 			& KNN 		& PKNN 		& MCMC 		& KOREA		\cr
			& 	 		&  			& (10000 runs) 	& 			\cr	
\hline
Crabs 		& 0.10		&0.46 		& 168.76		& 9.77		\cr
Fglass		& 0.11		&0.52			& 200.59		& 10.61		\cr
Meat 		& 0.12		&0.92			& 270.46		&15.66		\cr
Oliveoil		& 0.02		&0.13	 		& 34.58		& 1.90		\cr
Wine			& 0.08		&0.30			& 129.03		& 6.25		\cr
Iris			& 0.07		&0.26			& 95.47		& 5.14		\cr
\hline\hline
\end{tabular}
\end{table}

\section{Discussion}
\label{section: Discussion}

Our proposed algorithm uses an approach similar to the idea of INLA by replacing the model parameters with the model order (the number of 
neighbours, $k$). This means that we can speed up the computation by embedding (Quasi-)Newton method for Laplace approximation 
rather than grid sampling as done in the original INLA. However, as we can see in 
Fig. \ref{fig: Posterior distribution p(K, beta|Y)}, the posterior is not unimodal so we can find local optima rather than global 
optima for the maximal mode of the posterior if we use such a simple Laplace approximation. Therefore, instead of (Quasi-)Newton 
methods employed in the original INLA, we reconstructed the density with relatively slower grid approach for the real datasets in the PKNN of this paper. Of course, if the distribution is uni-modal, then we can use the Quasi-Newton method to speed up the algorithm.

\section{Conclusion}
\label{section: Conclusion}

We proposed a model selection algorithm for probabilistic k-nearest neighbour (PKNN) classification which is based on functional 
approximation in Bayesian framework. This algorithm has several advantages compared to other conventional model selection 
techniques. First of all, the proposed approach can quickly provide a proper distribution of the model order $k$ which is not 
given by other approaches, in contrast to time consuming techniques like MCMC. In addition, since 
the proposed algorithm is based on a Bayesian scheme, we do not need to run cross validation which is usually used for the 
performance evaluation. The proposed algorithm can also inherit the power of the fast functional approximation of INLA.  For 
instance, it can quickly find the optimal number of neighbours $k$ and efficiently generate the grid samples by embedding 
Quasi-Newton method if the posterior is uni-modal. Lastly, the proposed approach can  calculate the model average  which is 
marginalized posterior $p({\bf x}|{\cal Y})=\int_{{\cal M}}p({\bf x}|{\cal Y}, {\cal M})p({\cal M}|{\cal Y})d{\cal M}$.
We also remark that our algorithm is based on a pseudo-likelihood approximation of the likelihood and suggest that, although
our algorithm has yielded good performance, further improvements may result by utilising more accurate approximations of the 
likelihood, albeit at the expense of computational run time.

\subsubsection*{Acknowledgements}
This research was supported by the MKE (The Ministry of Knowledge Economy), Korea, under the ITRC (Information Technology Research Center) support program (NIPA-2012- H0301-12-3007) supervised by the NIPA (National IT Industry Promotion Agency). Nial Friel's research was supported by a Science Foundation Ireland Research Frontiers Program grant, 09/RFP/MTH2199.

\bibliographystyle{elsarticle-num}
\bibliography{refs}







\end{document}